\documentclass[preprint,12pt,authoryear]{elsarticle}

\usepackage{amssymb}
\usepackage{amsthm}

\usepackage{enumitem}
\usepackage{amsmath}
\usepackage{makecell}
\usepackage{multirow}
\usepackage{adjustbox}
\usepackage{booktabs}
\usepackage{graphicx}
\usepackage{float}
\usepackage{xcolor}
\usepackage{mathtools}
\usepackage{etoolbox}
\usepackage{setspace}
\usepackage{units}
\usepackage{subcaption}
\usepackage{comment}
\usepackage[ruled,vlined]{algorithm2e}
\usepackage[hidelinks]{hyperref}
\usepackage[margin=2.5cm]{geometry}

\definecolor{red(ncs)}{rgb}{0.0, 0.0, 0.0}

\definecolor{red(ncss)}{rgb}{0.0, 0.0, 0.0}
\newcommand{\rdd}[1]{\textcolor{red(ncss)}{#1}}

\definecolor{red(ncsss)}{rgb}{0.0, 0.0, 0.0}
\newcommand{\rrd}[1]{\textcolor{red(ncsss)}{#1}}

\newtheorem*{assumption*}{\assumptionnumber}
\providecommand{\assumptionnumber}{}
\makeatletter
\newenvironment{assumption}[2]
 {%
  \renewcommand{\assumptionnumber}{Assumption #1: #2}%
  \begin{assumption*}%
  \protected@edef\@currentlabel{#1}%
 }
 {%
  \end{assumption*}
 }

\DeclareMathOperator*{\argmax}{arg\,max}

\journal{}

\begin{document}

\begin{frontmatter}



\title{Two-step dynamic obstacle avoidance}


\author[1]{Fabian Hart}
\ead{fabian.hart@tu-dresden.de}

\author[1]{Martin Waltz\corref{CorrespondingAuthor}}
\ead{martin.waltz@tu-dresden.de}

\author[1,2]{Ostap Okhrin}
\ead{ostap.okhrin@tu-dresden.de}

\affiliation[1]{organization={Technische Universität Dresden, Chair of Econometrics and Statistics, esp. in the Transport Sector},
            city={Dresden},
            postcode={01062, Wuerzburger Str. 35}, 
            country={Germany}}
\affiliation[2]{organization={Center for Scalable Data Analytics and Artificial Intelligence (ScaDS.AI)}, city={Dresden/Leipzig}, country={Germany}}

\cortext[CorrespondingAuthor]{Corresponding author}
\begin{abstract}
Dynamic obstacle avoidance (DOA) is a fundamental challenge for any autonomous vehicle, independent of whether it operates in sea, air, or land. This paper proposes a two-step architecture for handling DOA tasks by combining supervised and reinforcement learning (RL). In the first step, we introduce a data-driven approach to estimate the collision risk \rdd{(CR)} of an obstacle using a recurrent neural network, which is trained in a supervised fashion and offers robustness to non-linear obstacle movements. In the second step, we include these \rdd{CR} estimates into the observation space of an RL agent to increase its situational awareness.~We illustrate the power of our two-step approach by training different RL agents in a challenging environment that requires to navigate amid multiple obstacles. The non-linear movements of obstacles are exemplarily modeled based on stochastic processes and periodic patterns, although our architecture is suitable for any obstacle dynamics. The experiments reveal that integrating our \rdd{CR} metrics into the observation space doubles the performance in terms of reward, which is equivalent to halving the number of collisions in the considered environment. \rdd{We also perform a generalization experiment to validate the proposal in an RL environment based on maritime traffic and real-world vessel trajectory data.} Furthermore, we show that the architecture's performance improvement is independent of the applied RL algorithm.
\end{abstract}




\end{frontmatter}



\section{Introduction}
\label{sec:introduction}
Autonomous vehicles represent a promising avenue for enhancing safety and optimizing energy and fuel efficiency within traffic systems, making them a vital element of future sustainable transportation networks \citep{feng2023dense}. While the general public often fixates on self-driving cars \citep{badue2021self}, the potential benefits of autonomous vehicles extend to other modes of transportation, such as autonomous vessels \citep{negenborn2023autonomous} and aircraft systems \citep{ribeiro2020review}. Regardless of whether a vehicle is under human or autonomous control, the paramount concern in all traffic operations remains ensuring safety by preventing collisions with other vehicles and stationary obstacles \citep{kuchar2000review, huang2020ship}.

We focus on entities operating in non-lane-based traffic environments, irrespective of whether they steer a vessel, fly an aircraft, command a drone, or control any other vehicle. These entities are referred to as \emph{agents}. Furthermore, we will call traffic participants close to the agent \emph{obstacles}. Generally, an agent's objective is to reach a specific destination safely. A real-world system typically relies on a perception-planning-control (PPC) modularization \citep{sauer2018conditional, fossen2021handbook}. The perception module includes sensor systems such as LiDAR, radar, or camera to localize the agent and relevant obstacles in the environment accurately \citep{liu2016unmanned}. Moreover, considering the noise and uncertainties associated with real-world sensors, advanced state estimation techniques are often applied \citep{kandepu2008applying}. The planning module processes the information from the perception module and generates a safe and feasible local path in the form of a set of waypoints \citep{marin2018global}. Crucially, the path planning should consider the maneuverability of both the agent and the obstacles. Lastly, the control module translates high-level commands derived from the local path into low-level actuator control commands \citep{vagale2021path}.

In this paper, we consider the task of dynamic obstacle avoidance (DOA, \citealt{falanga2020dynamic}), which we define as the challenge of executing real-time collision avoidance maneuvers while facing obstacles with non-zero velocity vectors. Following the formalism of \cite{ribeiro2020review}, we consider a distributed setting where no communication between traffic participants takes place, which is a practically occurring situation across traffic domains. DOA can be attributed to local path planning in the PPC modularization\rdd{, and we review popular methods for DOA in Section \ref{sec:related_work}.} In the following, we deviate from traditional methods and consider specifically the emerging method of deep reinforcement learning (DRL, \citealt{sutton2018reinforcement}). DRL tackles sequential decision tasks, in which an agent interacts with an environment and learns based on trial-and-error. Over the last years, the methodology has led to remarkable successes in various application domains \citep{vinyals2019grandmaster, bellemare2020autonomous, fawzi2022discovering, chen2023deep, yun2024doubly}. In our DOA task, the agent receives an observation of the traffic situation, takes a steering or acceleration action, and obtains a numerical reward signal that should be maximized.

Several contributions emphasize the strong dependence of the reinforcement learning \rdd{(RL)} algorithms' performances on the precise selection of the features of the observation vector \citep{mousavi2016deep, suarez2019feature, li2019reinforcement}. These features build the basis for the agent's decision-making and should contain relevant information to handle the task at hand successfully. In DOA tasks, the observation features frequently consist of the position and velocity of the agent and nearby obstacles \citep{everett2018motion, brittain2022scalable, everett2021collision}. In real-world situations, we acknowledge that certain rule sets, like the Convention on the International Regulations for Preventing Collisions at Sea, explicitly require estimating collision risk \rdd{(CR)} with other traffic participants (Rule 7, \citealt{COLREGs1972}). Leveraging these considerations, we aim to additionally include \rdd{CR} metrics into the observation of the DRL agent.

We emphasize that a similar approach has been pursued by some prior works \citep{chun2021deep, xu2022colregs, waltz20232}. However, these studies share the limitation of assuming a \emph{linear motion} of nearby obstacles during the CR computation. For instance, popular CR measures based on the closest point of approach (CPA, \citealt{lenart1983collision}) like the distance between the vehicles at the CPA, denoted $d^{\rm CPA}$, and the time until reaching the CPA, denoted $t^{\rm CPA}$, are calculated assuming constant course and speed of the obstacle. While these widely used metrics offer great interpretability and constitute an essential tool for real-world captains and pilots, future turns or accelerations of the obstacles are not considered, leaving a significant risk of misinterpreting a given scenario.

\rdd{Addressing this challenge, we propose to merge traditional highly interpretable CPA-based measures with a supervised learning approach to CR estimation to construct a comprehensive DRL agent for DOA tasks.} In particular, our contributions are as follows:
\begin{itemize}  
    \item We propose a \emph{data-driven} approach to generate an estimate of the $d^{\rm CPA}$ and $t^{\rm CPA}$ with obstacles. More precisely, we train an LSTM network \citep{hochreiter1997long} to estimate the future trajectories of obstacles in a supervised fashion \citep{lecun2015deep}, and compute the CR metrics subsequently. This procedure can generalize across situations and is robust to non-linear obstacle behavior.
    \item We augment the \emph{observation space} of an RL algorithm to include the newly estimated CR metrics, enabling the agent to assess and solve various encounter situations robustly. Moreover, this proposal can be seen in context to recent advances from the robotics domain \citep{lee2019making} since our CR estimator acts as an additional observation processing module.
    \item We showcase the potential of our \emph{two-step architecture} (Step 1: CR estimation, Step 2: DRL) for handling DOA tasks, thereby underscoring its suitability for implementation inside a local path planning unit of a PPC modularization. More precisely, we define a traffic-type independent environment where an agent needs to avoid dynamic obstacles and quantify the advantage in terms of the reward difference of the proposal in comparison to a 'conventional' DRL agent for obstacle avoidance.
    \item \rdd{Using real-world vessel trajectory data, we demonstrate that the proposed two-step architecture effectively adapts to this new environment, highlighting the architecture's robustness and generalizability. This study underscores the potential of our approach for diverse applications beyond the initial traffic-type independent environment.}
\end{itemize}

The paper is structured as follows: \rdd{Section \ref{sec:related_work} reviews related work, while} Section \ref{sec:problem} provides a concise and formal problem description of the considered DOA task. \rrd{Section \ref{sec:proposed_method} describes the newly proposed method, while Section \ref{sec:Implementation} provides additional implementation details.} Section \ref{sec:results} shows the \rrd{experiments}, and \rdd{Section \ref{sec:gen_study} displays the generalization study to maritime traffic.} \rdd{Section \ref{sec:discussion} discusses limitations and outlines areas for future work.} Section \ref{sec:conclusion} concludes this article. The source code for this paper is publicly available at \rdd{\url{https://github.com/MarWaltz/Two-Step-DOA}}.

\section{\rdd{Related work}}\label{sec:related_work}
\subsection{\rdd{Dynamic obstacle avoidance: Method overview}}
\rdd{DOA is a problem originating from the robotics literature with a longstanding history of seminal contributions; see \cite{reif1994motion}, \cite{latombe2012robot}. Key methods include the artificial potential field (APF) method introduced by \citet{khatib1986real}, which uses virtual forces to push robots away from obstacles. The vector field histogram method by \citet{borenstein1991vector} offers a fast and efficient way to navigate dynamic environments by creating a polar histogram of obstacle densities. Another significant contribution is the dynamic window approach (DWA) proposed by \citet{fox1997dynamic}, which integrates the robot's dynamics into the obstacle avoidance process, enabling more realistic and feasible maneuvers. Furthermore, the curvature-velocity method introduced by \citet{simmons1996curvature} solves a constrained optimization problem to explicitly trade off speed, safety, and goal-directness to enable fast and smooth robot navigation. \cite{fiorini1998motion} introduced velocity obstacles (VO), which defines a set of velocities that would result in a collision with an obstacle, allowing the robot to choose a velocity that avoids these dangerous trajectories. Further influential approaches to DOA for mobile robots include ant colony optimization \citep{dorigo1997ant}, sampling-based approaches like rapidly-exploring random trees \citep{lavalle2001rapidly}, or particle swarm optimization \citep{kennedy1995particle}.}

\rdd{Other traffic domains have adapted these methods to suit their individual needs. For instance, in the maritime literature, \cite{lyu2019colregs} and \cite{liu2023colregs} have adjusted the APF method for DOA of vessels, while \cite{kuwata2013safe} integrate maritime traffic rules into the VO approach by \cite{fiorini1998motion}. In addition, \cite{hu2023path} outline a path planning strategy for autonomous ships using the DWA. In the aviation community, \cite{tan2020three} use a three-dimensional VO approach to control unmanned aerial vehicles, and \cite{falanga2020dynamic} employ an APF approach to enable micro aerial vehicles to dodge fast-moving objects. For further applications and information on conventional DOA methods, we refer to the comprehensive reviews by \cite{kuchar2000review} and \cite{ribeiro2020review} for the aviation literature, \cite{polvara2018obstacle} and \cite{huang2020ship} for autonomous vessels, and \cite{pandey2017mobile} and \cite{patle2019review} for mobile robots.}

\subsection{\rdd{Dynamic obstacle avoidance: Reinforcement learning}}
\rdd{The method employed in this paper, DRL, has also been successfully applied to DOA tasks across traffic domains. For instance, \cite{zhao2021physics} propose a physics-informed DRL model for resolving aircraft conflicts, while \cite{brittain2022scalable} use a multi-agent RL approach to control aircraft in high-density en route airspaces. \cite{wang2019autonomous} introduce a recurrent DRL approach to autonomously navigate unmanned aerial vehicles (UAVs) in complex three-dimensional environments. Additionally, \cite{isufaj2022multi} apply DRL with graph neural networks to resolve conflicts among UAVs. In the maritime domain, \cite{WaltzOkhrin2022COLREG} and \cite{heiberg2022risk} develop DRL approaches that explicitly consider the CR posed by nearby vessels and adhere to maritime traffic rules. \cite{xu2022colregs} integrate a priority sampling mechanism into a DRL framework to facilitate vessel path planning while accounting for dynamic obstacles. In addition, \cite{shen2019automatic} demonstrate a real-world application of DRL to avoid collisions involving three self-propelled ships autonomously.}

\rdd{Concerning mobile robot navigation, \cite{everett2018motion} outline a recurrent DRL method to navigate up to 20 agents jointly in an environment while performing DOA. \cite{duguleana2016neural} combine $Q$-Learning \citep{watkins1992q} with a neural network-based trajectory planner to let a robot avoid obstacles in a virtual and a real-world environment. In addition, \cite{xue2019deep} use the Double DQN of \cite{van2016deep} for a mobile robot DOA task, which is evaluated on a physical robot platform. Further contributions to DOA via DRL in non-lane-based traffic scenarios include \cite{jiang2022human}, \cite{xu2020collision}, \cite{everett2018motion}, and \cite{roghair2022vision}.}

\subsection{\rdd{Collision risk estimation}}
\rdd{DOA involves assessing CR with nearby obstacles to determine appropriate avoidance maneuvers. Methods for quantifying CR in non-lane-based traffic vary widely, focusing on geometric considerations of current traffic states, defining safety zones, and uncertainty in future states; see the reviews of \cite{chen2019probabilistic} and \cite{ozturk2019individual}. Key contributions to the aviation literature include \cite{paielli1997conflict}, who estimate the probability of conflict (PC) between vehicles based on trajectory predictions and associated prediction error, assuming a Gaussian distribution. \cite{hwang2008intent} adapt \cite{paielli1997conflict} to better handle vehicle maneuvers and non-linear behaviors, while \cite{yang2004real} outline a Monte Carlo method for PC computation. More recently, \cite{zhang2020collision} propose a probabilistic model to evaluate CR between an intruding drone and commercial aircraft. \cite{zou2021collision} derive rapid methods for collision probability estimation for three types of collision zones involving small unmanned aircraft systems. Finally, \cite{mitici2018mathematical} present a unified mathematical framework for collision probability estimation in the air traffic literature.}

\rdd{In maritime contexts, \cite{fujii1971traffic} introduce a safe zone around vessels called ship domain, which is empirically investigated using real-world vessel data in \cite{hansen2013empirical}. Moreover, \cite{szlapczynski2017review} detail recent developments and discuss several modifications of the ship domain concept. In addition, \cite{mou2010study} quantify CR using an exponential relationship with the CPA metrics of \cite{lenart1983collision}, and \cite{debnath2010navigational} model the maximum of the CPA metrics of nearby vessels with a truncated gamma distribution. In recent years, \cite{zheng2020svm} build a support vector machine classifier to model the probability of a vessel collision while accounting for the consequences of a potential collision. \cite{tengesdal2021ship} estimate the ship collision probability leveraging the cross-entropy method \citep{kroese2006cross}, and integrate the estimates in the probabilistic sampling-based model predictive control approach of \cite{tengesdal2020risk} to perform a DOA task. Lastly, \cite{yu2022framework} outline a multi-criteria framework for real-time CR assessment that is validated in Portuguese continental coastal waters.}

\section{Problem description}
\label{sec:problem}
We consider a two-dimensional DOA task detached from a specific transportation context. We define a point-mass agent with a constant longitudinal speed for simplification while the lateral acceleration can be controlled. Since this study aims at a realistic setting with non-linearly moving obstacles, we generically define an obstacle trajectory as: 
\vspace{-0.3cm}
\begin{equation}\label{eq:obstacle_traj}
\begin{aligned}[c]
    \vec{p}_{t} &=  \vec{p}_{t, \rm lin} + \vec{p}_{t, \rm non},
\end{aligned}
\end{equation}
where $\vec{p}_t = (x_{t}, y_{t})^\top \in \mathbb{R}^2$ is the position of the obstacle at time step $t$. The terms $ \vec{p}_{t, \rm lin}$ and $\vec{p}_{t, \rm non}$ denote the linear and non-linear components of the obstacle trajectory, respectively. More precisely, the linear part is defined as:
\vspace{-0.3cm}
\begin{equation}\label{eq:obstacle_traj_lin}
\begin{aligned}[c]
     \vec{p}_{t, \rm lin} &= \vec{p}_0 + \Dot{\vec{p}}\, t,
\end{aligned}
\end{equation}
with the initial obstacle position $\vec{p}_0=(x_{0}, y_{0})^\top \in \mathbb{R}^2$ and the constant, obstacle-dependent velocity $\Dot{\vec{p}}=(\Dot{x}, \Dot{y})^\top \in \mathbb{R}^2$. Equation (\ref{eq:obstacle_traj_lin}) ensures that the obstacle moves in a specific direction. At the same time, the non-linear part $\vec{p}_{t, \rm non}$ can capture course deviations due to environmental effects or the obstacle's own avoidance maneuvers. Importantly, different practical behaviors of obstacles can be simulated via different specifications of $\vec{p}_{t, \rm non}$. Throughout this study, we exemplarily consider two different kinds of non-linear obstacle movements: \emph{stochastic} and \emph{periodic} trajectories.\\

\textbf{Stochastic trajectories.} The first non-linear specification is based on a stochastic two-dimensional auto-regressive process \citep{tsay2005analysis}:
\begin{equation}\label{eq:obstacle_traj_stoch_AR}
\begin{aligned}[c]
     A_{t+1} &= \phi_{\rm stoch} A_{t} + u_{t, \rm stoch}, \quad \text{where} \quad u_{t, \rm stoch} \sim \mathcal{N}_2\left[ \begin{pmatrix} 0\\ 0 \end{pmatrix},  \begin{pmatrix}\sigma_{\rm stoch}^2 & 0\\ 0 & \sigma_{\rm stoch}^2\end{pmatrix}\right],
\end{aligned}
\end{equation}
with initial value $A_{0} = (0,0)^\top$, auto-regressive parameter $\phi_{\rm stoch}$, and variance $\sigma_{\rm stoch}^2$ of the centered noise stemming from the multivariate normal distribution. Addressing the noisy nature of the process, we perform an exponential smoothing of (\ref{eq:obstacle_traj_stoch_AR}):
\begin{equation}\label{eq:obstacle_traj_AR_smooth}
\begin{aligned}[c]
	A_{t,\rm smooth} &=	\beta_{\rm stoch} A_{t}+(1-\beta_{\rm stoch}) A_{t-1,\rm smooth}, \quad \text{where} \quad 	A_{0,\rm smooth} = 0,
\end{aligned}
\end{equation}
with the smoothing factor $\beta_{\rm stoch} = 0.2$. Moreover, to account for different obstacle velocities, we scale the stochastic trajectory:
\begin{equation}\label{eq:obstacle_traj_non-lin_stoch}
\begin{aligned}[c]
    \vec{p}_{t, \rm non}^{\,(1)} = \frac{\| \Dot{\vec{p}} \|}{v_{\rm max}} A_{t,\rm smooth},
\end{aligned}
\end{equation}
with maximum obstacle speed $v_{\rm max}$.\\

\textbf{Periodic trajectories.} The second obstacle behavior considered in this study incorporates a sinus pattern with additional noise:
\begin{equation}\label{eq:obstacle_traj_sin}
\begin{aligned}[c]
    \vec{p}_{t, \rm non}^{\,(2)} = \begin{pmatrix} \Dot{y}\\ -\Dot{x}\end{pmatrix}  A_{\rm sin} \sin\left({\frac{2 \pi t}{T_{\rm sin}}}\right) + u_{t, \rm sin} , \quad \text{where} \quad u_{t, \rm sin} \sim \mathcal{N}_2\left[ \begin{pmatrix} 0\\ 0 \end{pmatrix},  \begin{pmatrix}\sigma_{\rm sin}^2 & 0\\ 0 & \sigma_{\rm sin}^2\end{pmatrix}\right],
\end{aligned}
\end{equation}
with amplitude $A_{\rm sin}$, period $T_{\rm sin}$, and variance $\sigma_{\rm sin}^2$.\\

Figures \ref{fig:AR1_prediction} and \ref{fig:sinus_prediction} visualize examples of the stochastic and periodic trajectories, respectively, in blue color. Throughout this study, we use a simulation step size of $\Delta t = \unit[5]{s}$ for simulating obstacle and agent movement. Crucially, to train a \emph{\rdd{CR} estimator} that estimates CR metrics for each obstacle and for each time step, we impose the following assumption:
\begin{assumption}{1}{Behavior stability}
The behavior of the obstacles used to train the collision risk estimator is similar to the obstacle behavior in the simulation environment.
\end{assumption}

\section{\rrd{Proposed method}}\label{sec:proposed_method}
\rrd{First, we provide an overview of the essential background in supervised and reinforcement learning in Sections \ref{subsec:SL_basics} and \ref{sec:RL_description}, respectively. On this basis, we present our proposed two-step method for DOA in Section \ref{sec:twostepmethod}.}

\subsection{\rrd{Supervised learning}}\label{subsec:SL_basics}
Supervised learning is the most common form of machine learning and describes the task of learning a mapping from potentially high-dimensional input samples to corresponding target vectors based on labeled data \citep{lecun2015deep, zhou2021survey}. Paired with (deep) neural networks as function approximators, supervised learning approaches have shown remarkable successes in various real-world problems such as speech recognition \citep{nassif2019speech} or object detection \citep{dhillon2020convolutional}.
In this study, we consider the regression task of having data pairs $D = \left\{(\mathbf{X}_i, \mathbf{Y}_i) \right\}_{i=1}^{N}$, where $\mathbf{X}_i \in \mathbb{R}^n$, $\mathbf{Y}_i \in \mathbb{R}^m$, and learning the parameter vector $\theta$ of a neural network $f_{\theta}: \mathbb{R}^n \rightarrow \mathbb{R}^m$ by minimizing the mean squared error $MSE(D;\theta) = N^{-1} \sum_{i=1}^{N}\left[f_{\theta}(\mathbf{X}_i) - \mathbf{Y}_i\right]^{\top}\left[f_{\theta}(\mathbf{X}_i) - \mathbf{Y}_i\right]$ with the Adam optimizer \citep{kingma2014adam}.

\subsection{\rrd{Reinforcement learning}}\label{sec:RL_description}
RL aims at solving sequential decision tasks in which an agent interacts with an environment under the objective to maximize a numerical reward signal \citep{sutton2018reinforcement}. Formally, we consider Markov Decision Processes (MDP, \citealt{puterman1994markov}) consisting of a state space $\mathcal{S}$, an action space $\mathcal{A}$, an initial state distribution $T_0: \mathcal{S} \rightarrow [0,1]$, a state transition probability distribution $\mathcal{P}: \mathcal{S} \times \mathcal{A} \times \mathcal{S} \rightarrow [0,1]$, a reward function $\mathcal{R}: \mathcal{S} \times \mathcal{A} \rightarrow \mathbb{R}$, and a discount factor $\gamma \in [0,1)$. At each time step $t$, the agent receives a state information $S_t \in \mathcal{S}$, selects an action $A_t \in \mathcal{A}$, gets an instantaneous reward $R_{t+1}$, and transitions based on the environmental dynamics $\mathcal{P}$ to the next state $S_{t+1} \in \mathcal{S}$. In practical applications, the full-state information is rarely available due to sensor limitations, delays, noise, or other disturbances. Addressing this issue, a generalization called Partially Observable Markov Decision Processes (POMDP, \citealt{kaelbling1998planning}) introduces two additional components: the observation space $\mathcal{O}$ and the observation function $\mathcal{Z}: \mathcal{S} \times \mathcal{A} \times \mathcal{O} \rightarrow [0,1]$. In a POMDP, the agent does not receive the new state $S_{t+1}$ directly, but instead an observation $O_{t+1} \in \mathcal{O}$, which is generated with probability $P(O_{t+1} | S_{t+1}, A_t)$ by the observation function $\mathcal{Z}$. Consequently, a POMDP is a Hidden Markov Model with actions and the observations are used for learning. In the following, we use capital notation, e.g., $S_t$, to indicate random variables and small notation, e.g., $s_t$ or $s$, to describe their realizations.

Objective of the agent in the MDP scenario is to optimize for a policy $\pi: \mathcal{S} \times \mathcal{A} \rightarrow [0,1]$, a mapping from states to probability distributions over actions, that maximizes the expected return, which is the expected discounted cumulative reward, from the start state: $E_{\pi}\left[ \sum_{k=0}^{\infty} \gamma^k R_{k+1}\right | S_0]$. Common practice is the definition of action value functions $Q^{\pi}(s,a)$, which are the expected return when starting in state $s$, taking action $a$, and following policy $\pi$ afterwards: $Q^{\pi}(s,a) = E_{\pi}\left[ \sum_{k=0}^{\infty} \gamma^k R_{t+k+1} | S_t = s, A_t = a \right]$. Crucially, in an MDP there is a deterministic optimal policy $\pi^*(s) = \argmax_{a \in \mathcal{A}} Q^*(s,a)$ if the state space is finite or countably infinite \citep{puterman1994markov}. The policy $\pi^*$ is connected with optimal action-values $Q^*(s,a) = \max_{\pi} Q^{\pi}(s,a)$. The general control objective of RL is to find or approximate $\pi^*$.

\subsection{\rrd{Two-step method}}\label{sec:twostepmethod}
\begin{figure}[ht!]
    \centering
    \includegraphics[width=\textwidth]{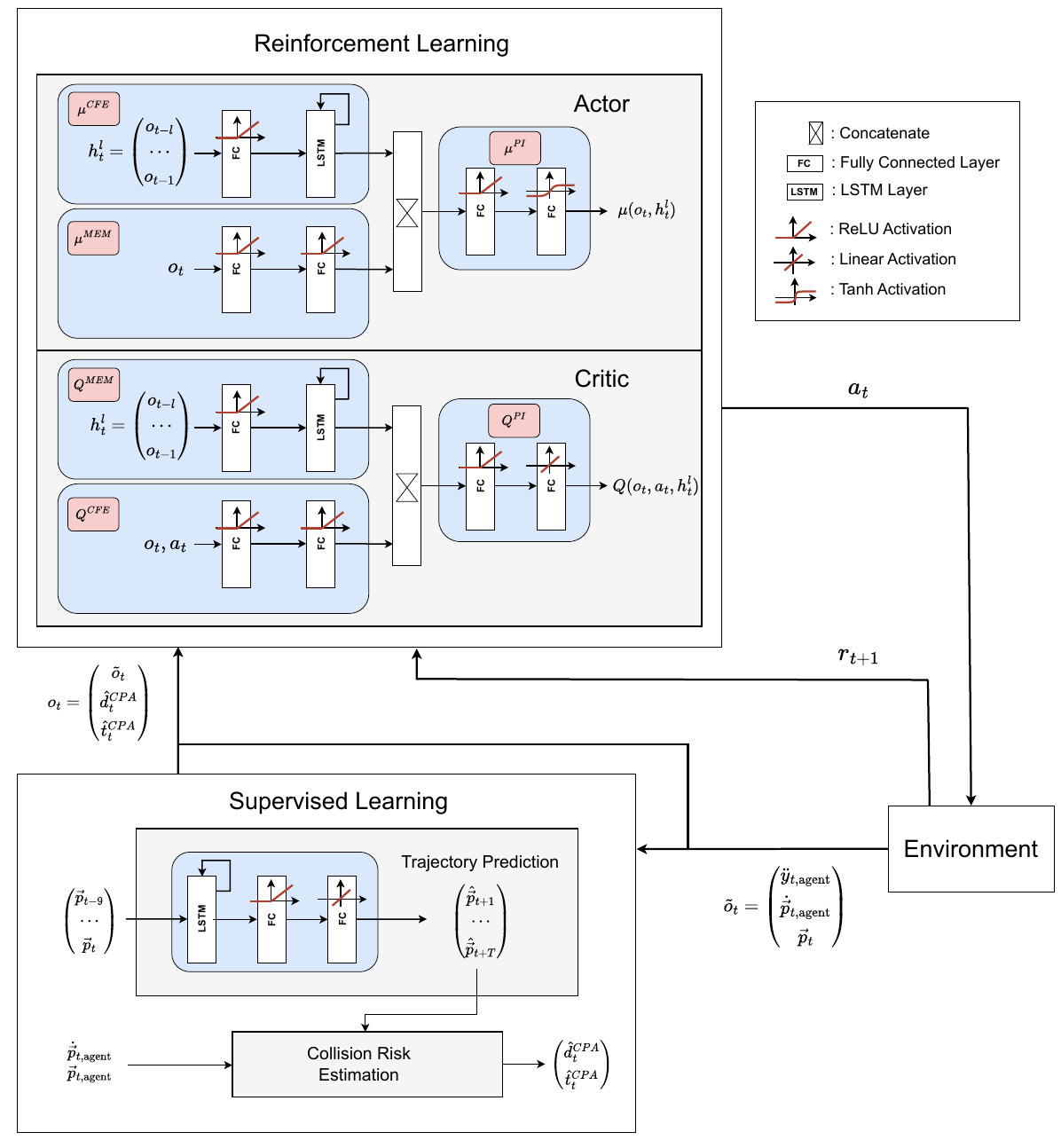}
	\caption[The proposed two-level architecture for DOA tasks]{The proposed two-level architecture for DOA tasks.}
	\label{fig:Architecture}
\end{figure}
\rrd{We visualize the proposed two-step method for DOA in Figure \ref{fig:Architecture}.~The approach is composed of two distinct steps, each contributing to the overall decision-making process. In the first step, a supervised learning module predicts the future trajectories of relevant obstacles while accounting for possible non-linear motions. These trajectory predictions are used to generate estimates of the $d^{\rm CPA}$ and $t^{\rm CPA}$, which are informative CR metrics that quantify the criticality of specific obstacles. In the second step, these CR metrics are concatenated with the remaining features from the RL environment to build a comprehensive observation vector for the RL agent. The agent iteratively interacts with the environment, adjusting its actions to maximize reward and optimize the DOA policy. Through this process, the RL agent learns to make informed, context-sensitive decisions that enhance the safety and efficiency of operations within dynamic environments.}

\section{\rrd{Implementation details}}\label{sec:Implementation}
\rrd{This section provides details on the implementation of the method outlined in Section \ref{sec:proposed_method} to solve the problem described in Section \ref{sec:problem}.}

\subsection{Trajectory prediction}\label{sec:trajectory_prediction}
\begin{figure}[ht!]
    \centering
    \includegraphics[width=0.9\textwidth]{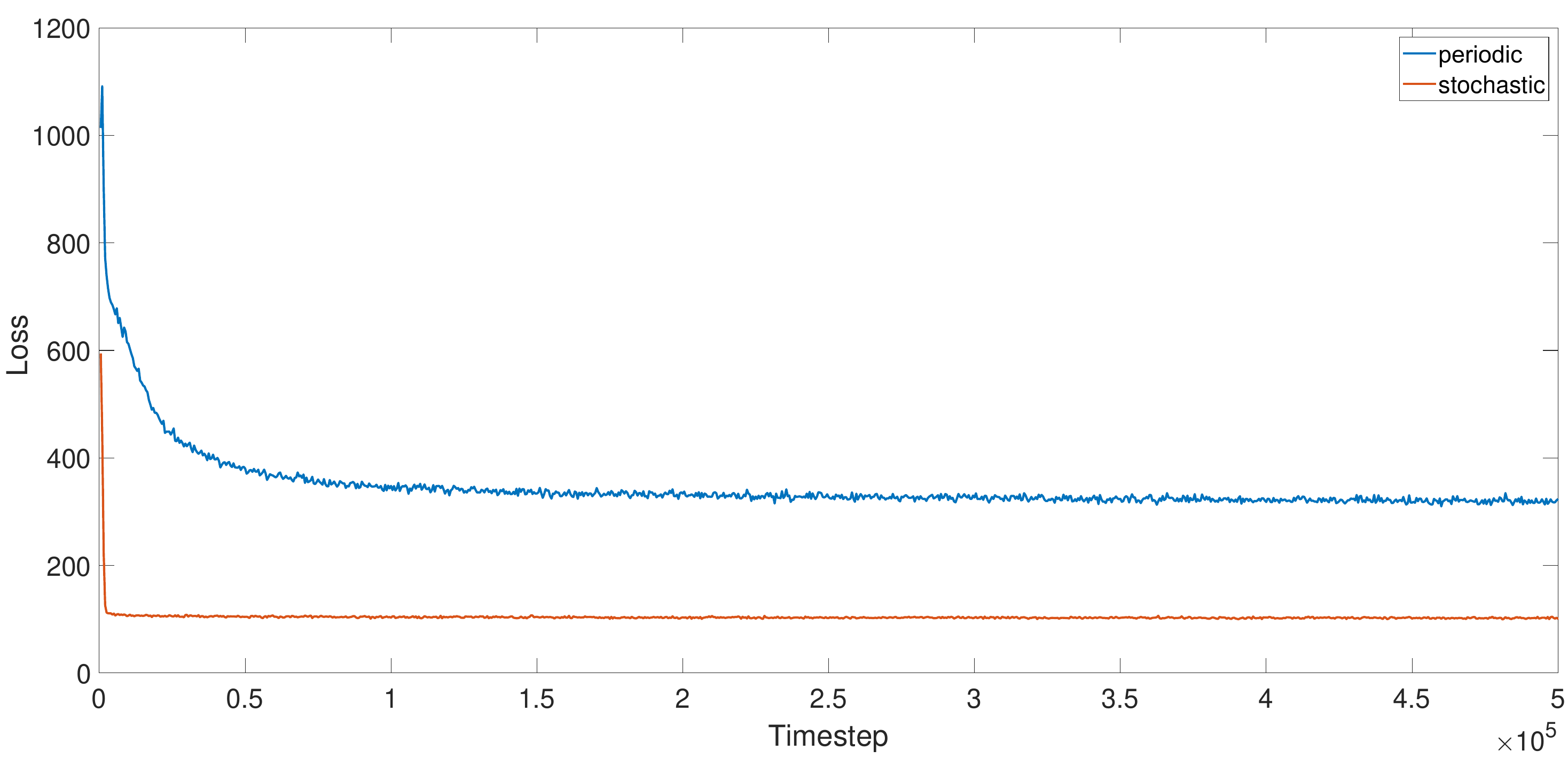}
	\caption[Training of the trajectory prediction module]{Training of the trajectory prediction module for periodic and stochastic obstacle trajectories.}
	\label{fig:estimator_training}
\end{figure}

The input features $\mathbf{X}$ \rrd{of the supervised learning module} contain information about $h=10$ previous positions of the obstacle $(\vec{p}_{t-9}, \dots, \vec{p}_t)^\top$, from which the neural network should approximate the next position $\vec{p}_{t+1}$. \rdd{However, we emphasize that we build the first differences of $(\vec{p}_{t-9}, \dots, \vec{p}_t)^\top$ before inputting them into the network, as we have empirically found that estimating relative rather than absolute positions is significantly more stable.} We construct the neural network $f_{\theta}$ with an LSTM-layer \citep{hochreiter1997long} of 64 hidden units. The output is forwarded through \rdd{two linear layers} with 64 neurons to yield output in $\mathbb{R}^2$. \rdd{To generate training data, we repeatedly sample trajectories according to the obstacle dynamics as given in (\ref{eq:obstacle_traj_non-lin_stoch}) and (\ref{eq:obstacle_traj_sin}) for the stochastic and periodic obstacle movements, respectively.} The training progress of the network is depicted in Figure~\ref{fig:estimator_training} for both defined non-linear obstacle movements. 

\begin{figure}[ht!]
	\centering
	\includegraphics[width=1\linewidth]{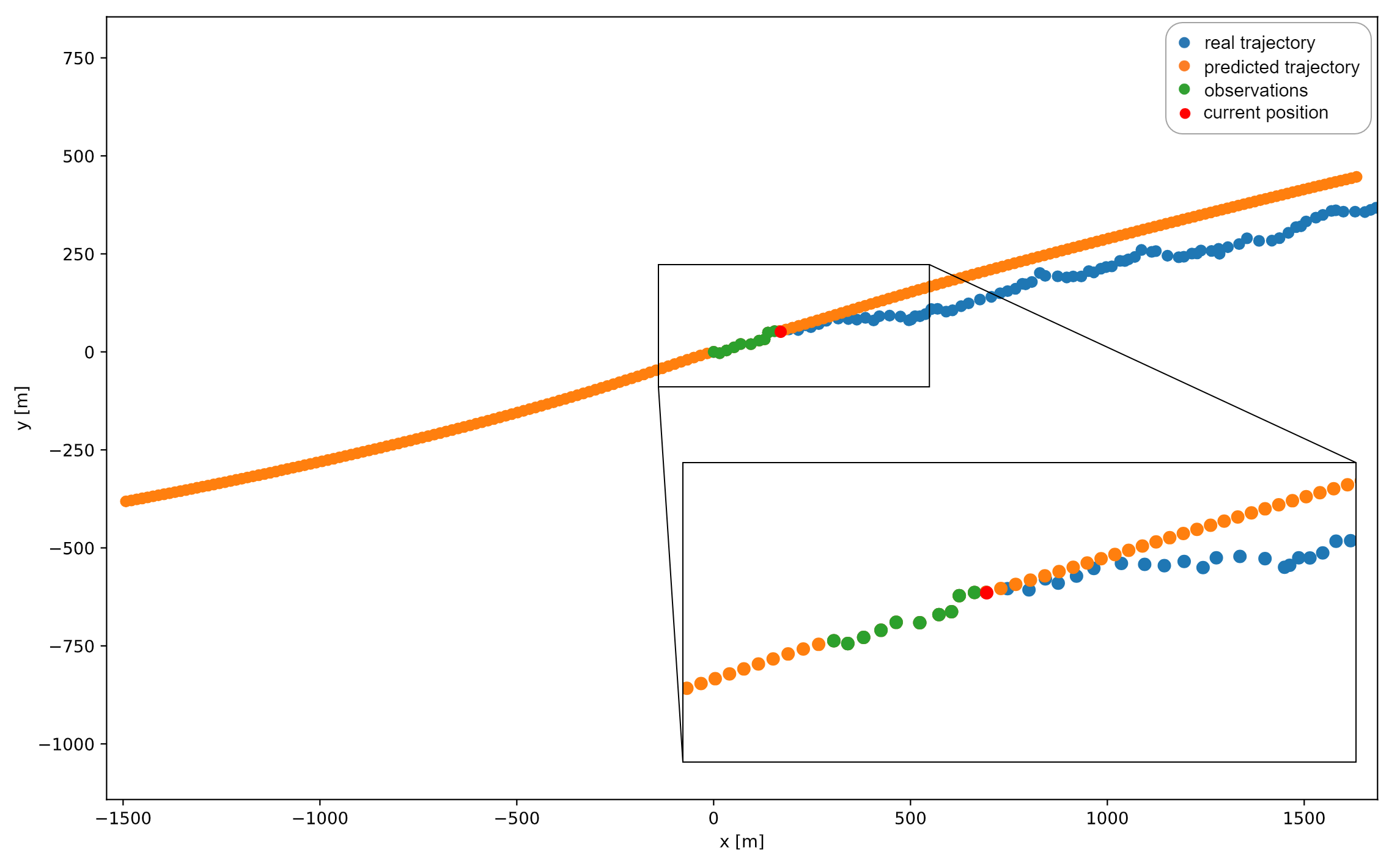}
	\caption[Trajectory prediction for stochastic obstacle movement]{Trajectory prediction for stochastic obstacle movement with last observations in green and ground truth data in blue.}
	\label{fig:AR1_prediction}
\end{figure}

\begin{figure}[ht!]
	\centering
	\includegraphics[width=1\linewidth]{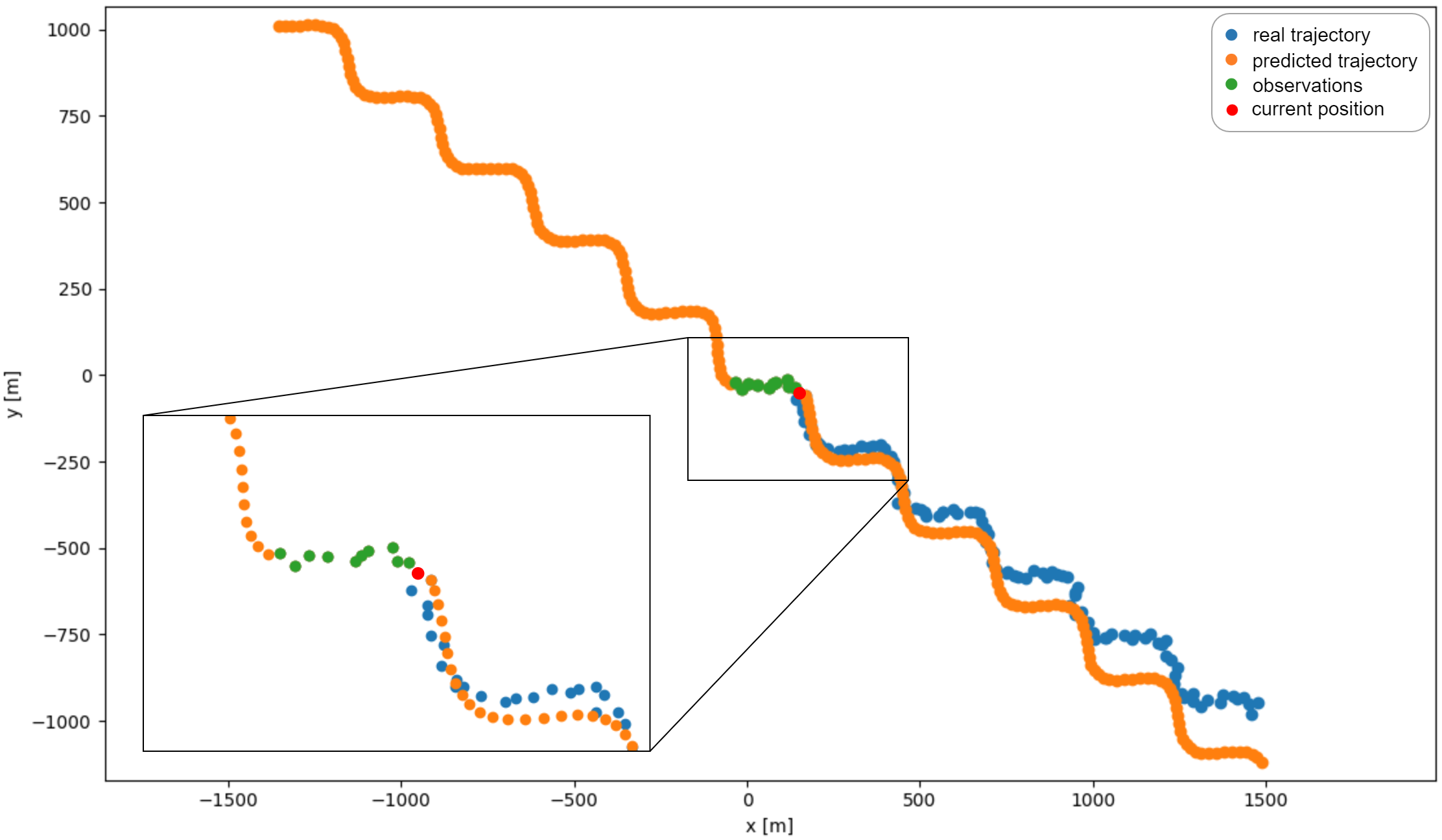}
	\caption[Trajectory prediction for periodic obstacle movement]{Trajectory prediction for periodic obstacle movement with last observations in green and ground truth data in blue.}
	\label{fig:sinus_prediction}
\end{figure}

After the training is completed, we perform subsequent one-step-ahead forecasts to predict an obstacle's \emph{entire trajectory} for a time horizon $T=100$. More precisely, we always take the last ten positions, possibly including predicted ones, to predict the following position iteratively. To illustrate the training results, we provide a representative example for a predicted trajectory in Figure \ref{fig:AR1_prediction} for the stochastic obstacle movement and Figure  \ref{fig:sinus_prediction} for the periodic movement. The blue dots represent the ground truth trajectory, from which only the first ten dots, shown in green, are observed. These observations are used to iteratively compute the predicted trajectory, which is shown in orange.
Since we assume that not all previous positions of an obstacle are observable, we predict past positions as well. 
These past obstacle positions are later used to estimate if the agent already passed and obstacle; see Section \ref{sec:CR_estimation}.
Since the obstacle trajectories defined in (\ref{eq:obstacle_traj_non-lin_stoch}) and (\ref{eq:obstacle_traj_sin}) are symmetric, we can use the same model to predict past positions.

\begin{figure}[ht!]
    \centering
    \includegraphics[width=\textwidth]{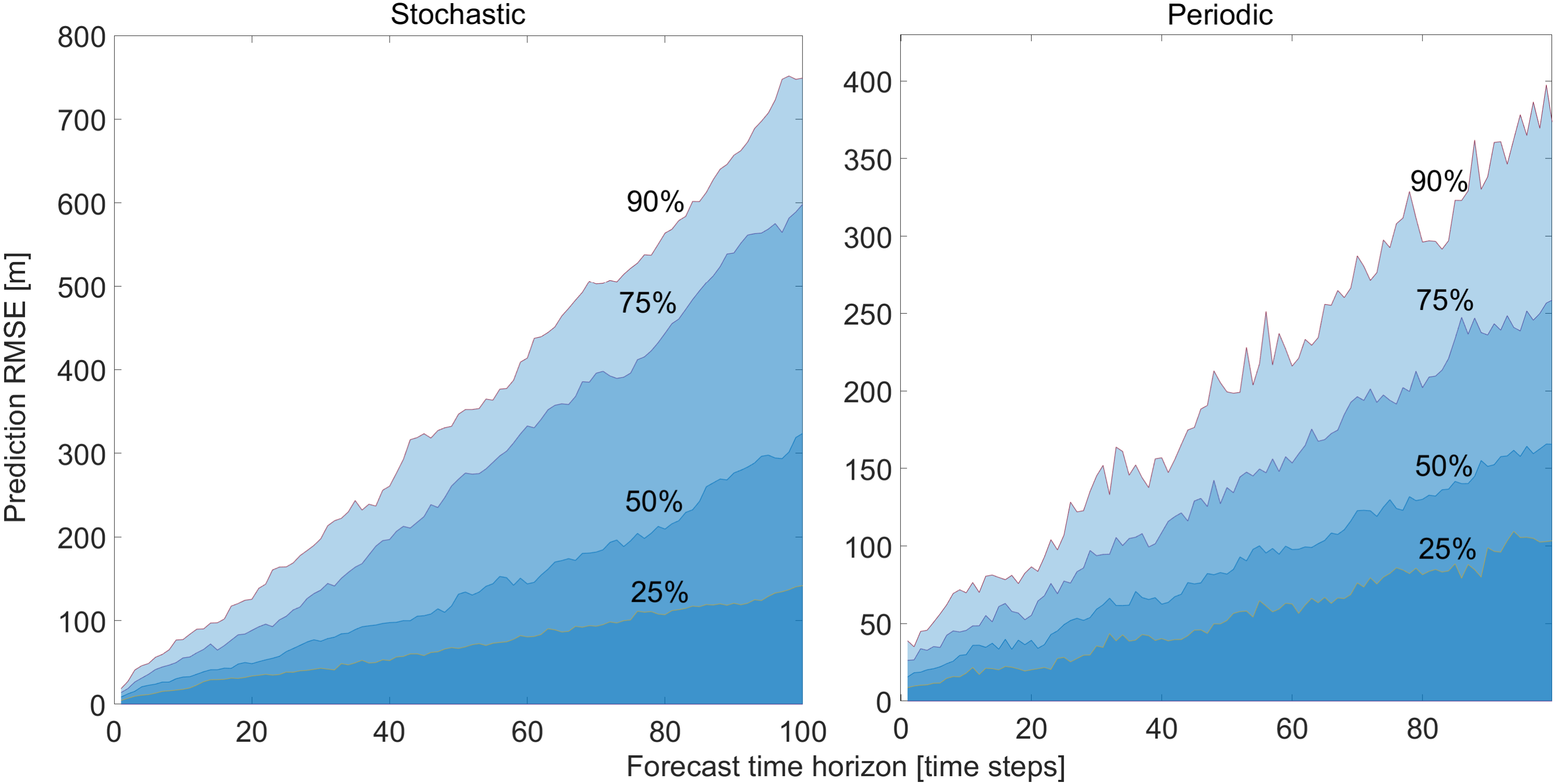}
	\caption[RMSE between predicted trajectory and real trajectory]{RMSE between predicted trajectory and real trajectory with 25\%-, 50\%-, 75\%-, and 90\%-quantile.}
	\label{fig:prediction_rmse}
\end{figure}

To quantitatively evaluate the prediction accuracy, we compute the root mean squared error \rdd{(RMSE)} between predicted and real trajectories for both obstacle movements, as shown in Figure \ref{fig:prediction_rmse}, with different quantiles. We can see that the prediction accuracy is relatively high for short-term predictions, but expectedly increases with increased forecasting time.

\subsection{Collision risk estimation}\label{sec:CR_estimation}
\begin{figure}[ht]
	\centering
	\includegraphics[width=1\linewidth]{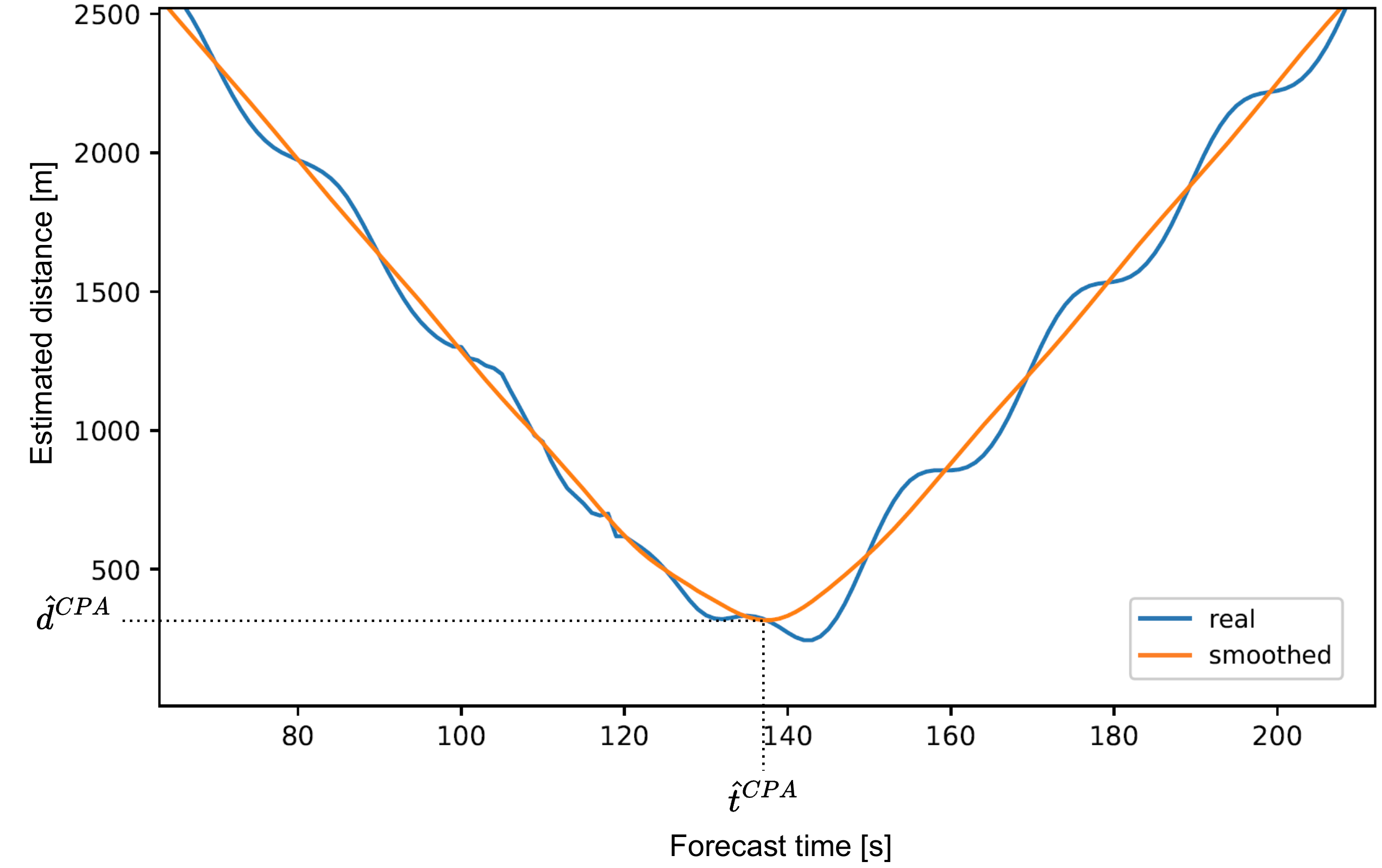}
	\caption[Real and smoothed distance between a predicted obstacle and predicted agent trajectory for a periodic obstacle movement]{Real and smoothed distance between a predicted obstacle and predicted agent trajectory for a periodic obstacle movement.}
	\label{fig:smoothing_distance}
\end{figure}

After predicting the trajectory of an obstacle for a time horizon $T = 100$, which equals a forecast time of $\unit[500]{s}$ based on a simulation step size of $\Delta t = \unit[5]{s}$, we can estimate the CPA by numerically finding the minimum between the predicted obstacle and the predicted agent trajectory. For the agent, we thereby assume a linear movement based on its current velocity vector. An example is illustrated in Figure \ref{fig:smoothing_distance} for a periodic movement of the obstacle, where the blue curve represents the estimated distance between the agent and obstacle based on these predictions. Naturally, these distance estimates exhibit the periodic oscillations of the obstacle trajectory. To reduce the variance in the estimation of $d^{\rm CPA}$ and $t^{\rm CPA}$ caused by this pattern, we apply a symmetric moving average filter on the estimated distance:
\begin{equation*}
\hat{d}_{j, \rm smooth} = \frac{1}{2n+1}\sum_{j=-n}^{n} \hat{d}_{j},
\end{equation*}
where $n=10$. The variables $\hat{d}_{j}$ and $\hat{d}_{j, \rm smooth}$ denote the distance estimates between agent and obstacle at step $j$ before and after smoothing, respectively. The resulting estimates for $d^{\rm CPA}$ and $t^{\rm CPA}$ using the smoothed distance are depicted in orange color in Figure \ref{fig:smoothing_distance}. Note that negative values for the $t^{\rm CPA}$ are possible if the agent already passed the obstacle.

\subsection{Recurrent reinforcement learning}\label{subsec:recurrentRL}
As described in Section \rrd{\ref{sec:RL_description}}, only observations $o_t$ rather than full states $s_t$ are available in the POMDP case. One popular approach to handle this scenario is the construction of belief states, which are distributions over the real states the agent might be in, given the observation so far. This approach requires a model of the environment and is computationally demanding \citep{heess2015memory}. An alternative approach might be to stack past observations together \citep{mnih2015human}, but this quickly increases the input dimension and limits the agent to a fixed number of frames. Finally, a further approach is to incorporate recurrency into the function approximators of model-free algorithms, which was shown to yield strong performances \citep{ni2021recurrent}. The recurrency enriches the agent's decision making by extracting information from past observations, potentially yielding an improved ability to solve problems without access to the complete state vector. 

In this study, we follow the latter approach and use the LSTM-TD3 algorithm of \cite{meng2021memory}, which adds LSTM layers to the off-policy, actor-critic TD3 algorithm of \cite{fujimoto2018addressing}. Crucially, the LSTM-TD3 uses a deterministic policy $\mu$ and defines the past history $h_{t}^{l}$ of length $l$ at time step $t$ defined as:
\begin{equation*}
h_t^l = \begin{cases} 
      o_{t-l}, \ldots, o_{t-1} & \text{if \hspace{0.125cm}} l, t \geq 1. \\
      o_0 & \text{else}. 
   \end{cases}
\end{equation*}
The zero-valued dummy observation $o_0$ is of the same dimension as a regular observation. Note that the definition of $h_{t}^{l}$ slightly differs from \cite{meng2021memory} since we do not include past actions in the history. Furthermore, we set $l = 10$ throughout the paper, analogous to the history length of the trajectory prediction from Section \rrd{\ref{sec:trajectory_prediction}}. The algorithm disassembles both actor and critic into different sub-components. Precisely, there is a memory extraction (MEM) part in the function approximators, $Q^{\rm MEM}$ and $\mu^{\rm MEM}$, respectively, that processes the history. In parallel, the current feature extraction (CFE) components $Q^{\rm CFE}$ and $\mu^{\rm CFE}$ process the observation of the current step $o_t$. Finally, the output of both MEM and CFE are concatenated and fed into the perception integration (PI) components $Q^{\rm PI}$ and $\mu^{\rm PI}$. These aggregate the extracted pieces of information and yield the final result. The network design of our LSTM-TD3 implementation is included in Figure \ref{fig:Architecture} and formalized as follows:
\begin{align*}
    Q(o_t, a_t, h_{t}^{l}) &= Q^{\rm PI}\left\{Q^{\rm MEM}(h_{t}^{l}) \bowtie Q^{\rm CFE}(o_t, a_t) \right\},\\
    \mu(o_t, h_{t}^{l}) &= \mu^{\rm PI}\left\{\mu^{\rm MEM}(h_{t}^{l}) \bowtie \mu^{\rm CFE}(o_t) \right\},
\end{align*}
where $\bowtie$ is the concatenation operator. The optimization of the algorithm is identical to the TD3. Moreover, to ensure the robustness of our architecture with respect to the algorithm, we apply the SAC algorithm of \cite{haarnoja2018soft, haarnoja2018soft2}, which uses a stochastic actor and automated tuning of the algorithm's temperature parameter. We construct the network design analogous to the LSTM-TD3 networks and refer to the approach in the following as LSTM-SAC.

\subsection{Training environment} \label{subsec:RL_env}
We aim to develop a challenging \rrd{RL} training environment for DOA where the agent has to anticipate several obstacle trajectories to avoid collisions. To solely focus on the collision avoidance task and \rdd{ensure the agent does not leave the environmental area of interest}, we define an environment where obstacles should be passed in a predefined fashion, so-called \emph{passing rules}. Moreover, research on RL-based DOA indicates that the final performance increases when the \rdd{CR} in training is high \citep{hart2022enhanced}, which we consider appropriately. 

We consider a set of obstacles $\mathcal{M} = \{1, \ldots, N_{\rm obstacle}\}$, where $N_{\rm obstacle} = 10$ is the total number of obstacles in the environment and the obstacle dynamics follow \eqref{eq:obstacle_traj}. 
We further define an agent with position  $\vec{p}_{t, \rm agent} = (x_{t,\rm agent}, y_{t,\rm agent})^\top $ and velocity $\Dot{\vec{p}}_{t, \rm agent}= (\Dot{x}_{\rm agent}, \Dot{y}_{t,\rm agent})^\top $ at time step $t$, where $\Dot{x}_{\rm agent}$ denotes a constant longitudinal velocity component. The RL agent controls the lateral velocity component via the lateral acceleration $\Ddot{y}_{t,\rm agent}$. Formally, the agent computes an action $a_t \in [-1,1]$ each time step $t$ that is mapped to the lateral acceleration:
\begin{equation*}
	\Ddot{y}_{t,\rm agent} = a_{y,\rm max} a_t,
\end{equation*}
where $a_{y,\rm max}$ defines the maximal lateral acceleration for the agent. The agent receives an observation vector each time step $t$, defined as:
\begin{equation}\label{eq:o_t}
\renewcommand*{\arraystretch}{1.5}
	o_{t} = \begin{pmatrix}
		\frac{\Ddot{y}_{t,\rm agent}}{a_{y,\rm max}},
		\frac{\Dot{\vec{p}}_{t,\rm agent}}{v_{\rm max}},
		\frac{\Vec{p}_{t,\rm agent}-\Vec{p}_{t,i}}{p_{\rm scale}},
        \frac{\hat{d}^{\rm CPA}_{t,i}}{d^{\rm CPA}_{\rm scale}},
        \frac{\hat{t}^{\rm CPA}_{t,i}}{t^{\rm CPA}_{\rm scale}}
        \end{pmatrix}^\top,
\end{equation}
where $v_{\rm max}$ is the maximum speed and $\Vec{p}_{t,i}$, $\hat{d}^{\rm CPA}_{t,i}$, and $\hat{t}^{\rm CPA}_{t,i}$ denote the position, estimated $d^{\rm CPA}$, and estimated $t^{\rm CPA}$ of the $i$-th obstacle, respectively. Crucially, the estimates $\hat{d}^{\rm CPA}_{t,i}$ and $\hat{t}^{\rm CPA}_{t,i}$ are thereby computed using the supervised learning module. The constants $p_{\rm scale}$, $d^{\rm CPA}_{\rm scale}$, and $t^{\rm CPA}_{\rm scale}$ are used for scaling. Since the computation of $\hat{d}^{\rm CPA}_{t,i}$ and $\hat{t}^{\rm CPA}_{t,i}$ is demanding, we update those values only every 10th simulation step. We emphasize that (\ref{eq:o_t}) slightly abuses notation and contains information about all obstacles $i \in \mathcal{M}$. Consequently, $o_t$ is of dimension $3 + 4 N_{\text{obstacle}}$.

The passing rule for each obstacle is indicated by the index within the observation vector. The first three observations are reserved for agent-related observations, the following $4N_{\rm obstacle}/2$ observations are reserved for obstacles with passing rule 'right', and the last $4N_{\rm obstacle}/2$ for obstacles with passing rule 'left'. Within these blocks of the same passing rule, obstacles are sorted according to ascending estimated $\hat{t}^{\rm CPA}$. To validate the proposed approach, we additionally configure a baseline agent with a reduced observation space, neglecting $\hat{d}^{\rm CPA}_{t,i}$ and $\hat{t}^{\rm CPA}_{t,i}$, and with sorting the obstacles according to ascending euclidean distance to the agent. 

Following recent RL literature \citep{silver2018general}, we aim to keep the reward function as simple as possible. Therefore, we penalize the agent only when passing an obstacle on the wrong side:

\begin{equation}\label{eq:reward_fnc}
	r_{t+1} = 
	\begin{cases}
	     -1, & \text{passing an obstacle contrary to the respective passing rule,}\\
		0 & \text{else.}
	\end{cases}
\end{equation}

\begin{figure}
    \centering
    \includegraphics[width=1\linewidth]{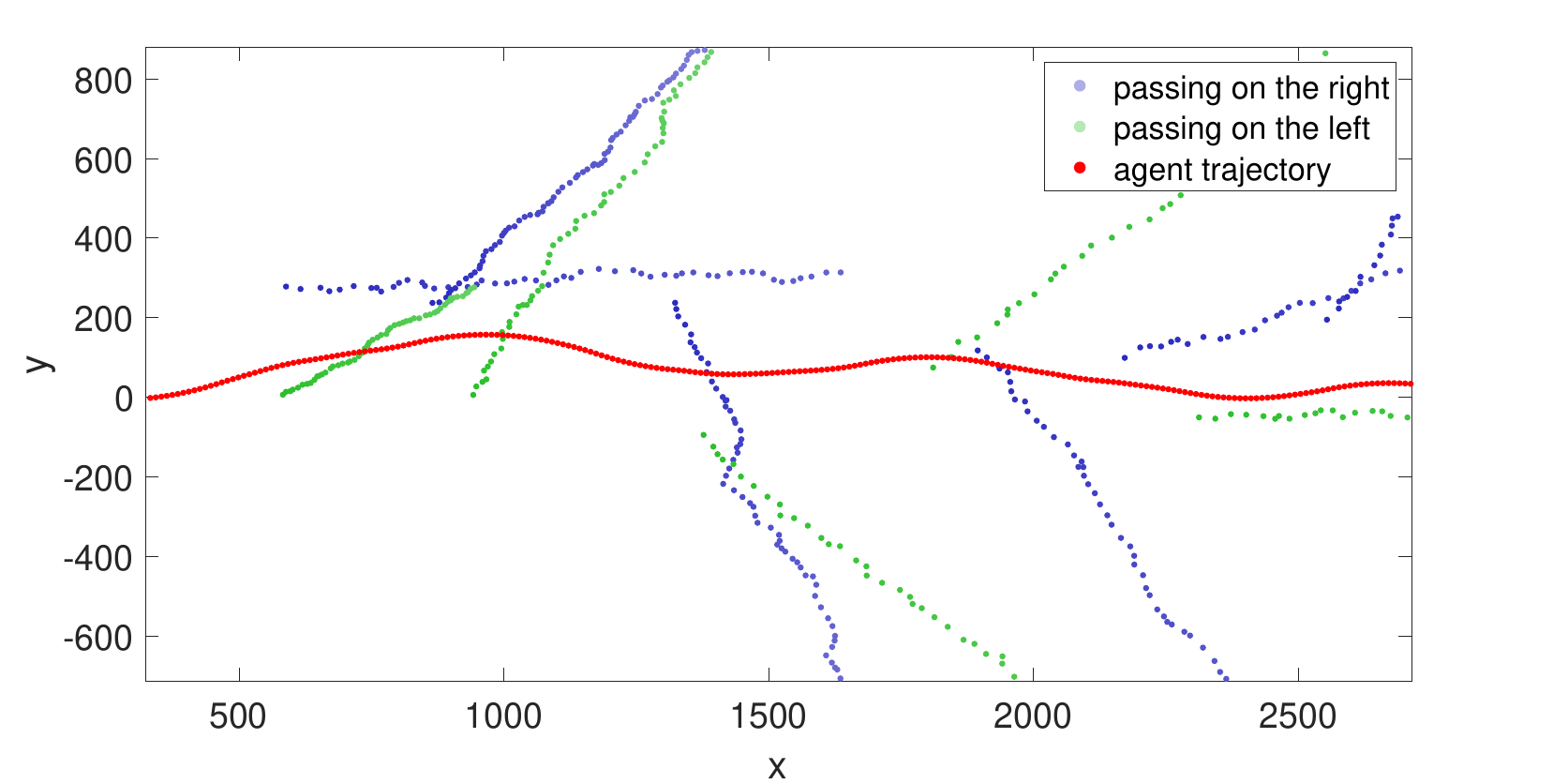}
    \caption[Training environment]{Training environment with several obstacle trajectories (green and blue) and a possible collision-free agent trajectory (red).}
    \label{fig:env_example}
\end{figure}

\ref{appendix:env_details} provides details about the initialization of the agent and the obstacles and other environment configurations. Figure \ref{fig:env_example} exemplary depicts a time interval of a training episode, where blue and green colors refer to obstacles with different passing rules. 

\section{\rrd{Experiments}}
\label{sec:results}
\subsection{\rdd{Setup}}
\begin{figure}[ht!]
	\centering
	\includegraphics[width=1\linewidth]{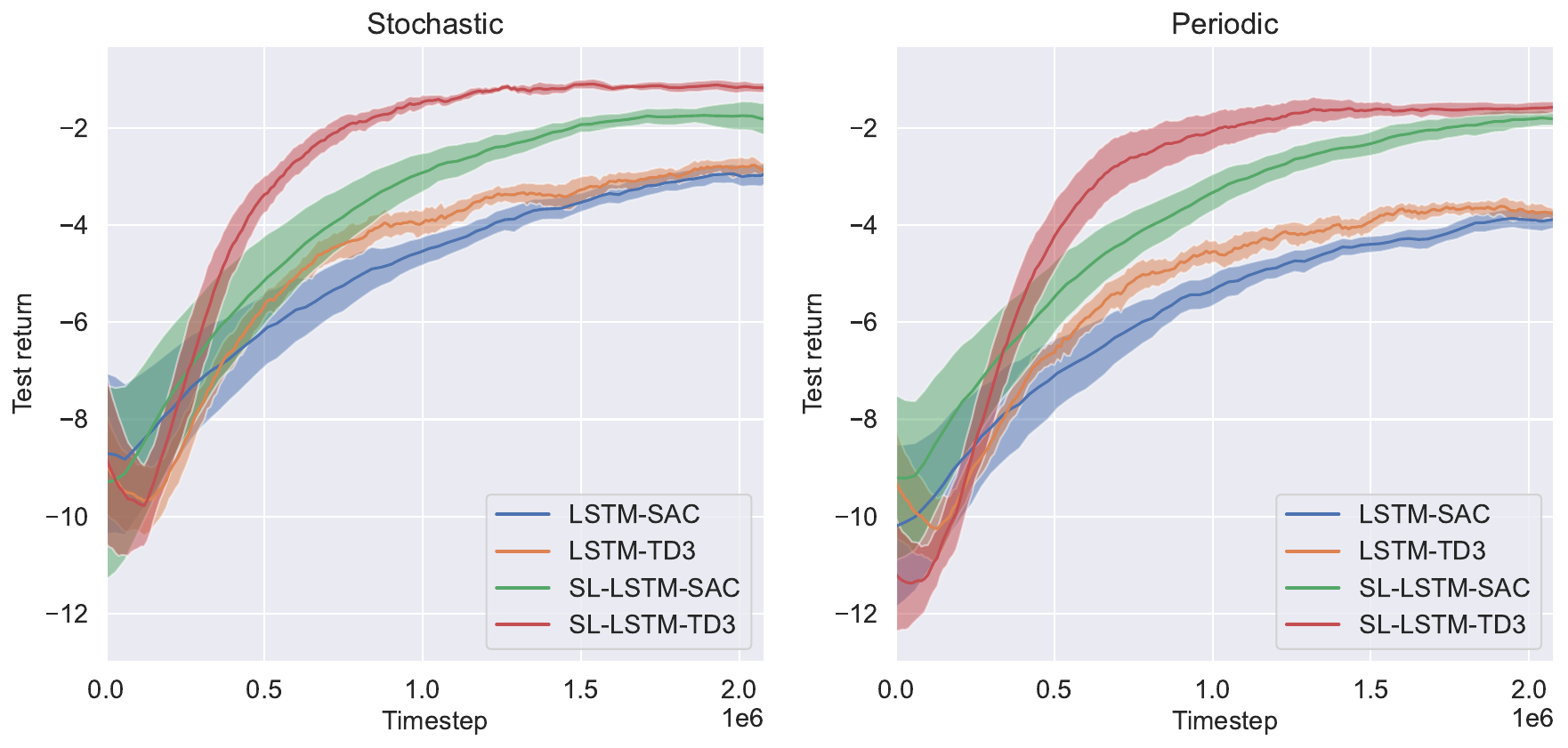}
	\caption[Performance comparison of agents]{\rdd{Performance comparison of agents that use the estimates for $d^{\rm CPA}$ and $t^{\rm CPA}$ (prefix 'SL') with baseline agents without the SL subroutine for stochastic (left) and periodic (right) obstacle movement. The y-axis displays the test return during an episode and the history length is $l=4$.}}
	\label{fig:trainResults}
\end{figure}

\begin{figure}[ht!]
    \centering
    \includegraphics[width=\linewidth]{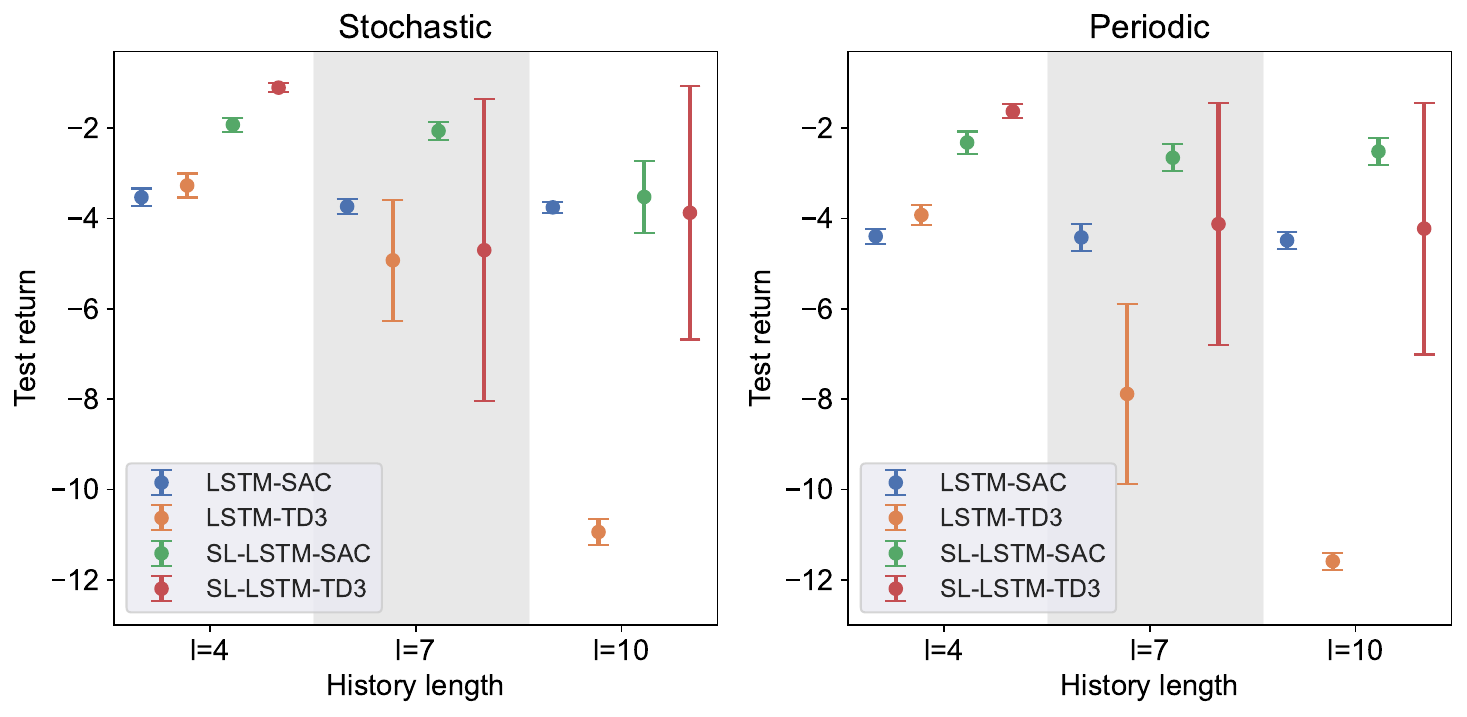}
    \caption[Ablation study for history length]{\rdd{Ablation study over the history lengths $l \in \{4,7,10\}$. The plots show the test return for the two obstacle movements after $1.5\cdot 10^6$ training steps, including a 95\% confidence interval based on 10 independent runs.}}\label{fig:h_ablate}
\end{figure}

We compare our proposed two-step architecture with a training procedure without incorporating the estimates of $d^{\rm CPA}$ and $t^{\rm CPA}$ in the observation vector. Crucially, to demonstrate the proposal's invariance to the selected DRL algorithm, we run the same experiments on two different actor-critic algorithms: the LSTM-TD3 of \cite{meng2021memory}, which uses a deterministic actor, and the LSTM-SAC as described in Section \rrd{\ref{subsec:recurrentRL}}, which uses a stochastic actor. All experiments are conducted in the same training environment, which is described in Section \ref{subsec:RL_env}. Hyperparameters shared by both algorithms are set equal; see Table \ref{tab:hyperparams_RL}. Figure \ref{fig:trainResults} displays the training results \rdd{using a history length of $l=4$,} where we mark agents trained with CR estimates from the supervised learning module with the prefix 'SL'. The results are averaged over ten independent runs of each experiment, and we include point-wise \rdd{95\%} confidence intervals for robustness. \rdd{In addition, we include an ablation study over the history length, which is shown in Figure \ref{fig:h_ablate}.} 

\subsection{\rdd{Analysis}}
Following (\ref{eq:reward_fnc}), the absolute value of the test return of an episode can be immediately interpreted as the number of collisions. Thus, \rdd{according to Figure \ref{fig:trainResults},} including the \rdd{CR} estimates based on supervised learning roughly halves the number of collisions. This performance boost is independent of the used algorithm\rdd{, and the variance of the test returns is relatively small for $l = 4$. Analyzing Figure \ref{fig:h_ablate}, we see that the results for larger history lengths underline the finding from $l=4$ that including the SL module is highly beneficial. In particular, for $l=10$, the LSTM-TD3 without the SL component fails to learn anything, while the SL-LSTM-TD3 approach achieves much better performance, although at the expense of increased variance. We generally observe that the LSTM-TD3 has a higher return variance than the LSTM-SAC.}

\begin{figure}[ht!]
	\centering
	\includegraphics[width=1\linewidth]{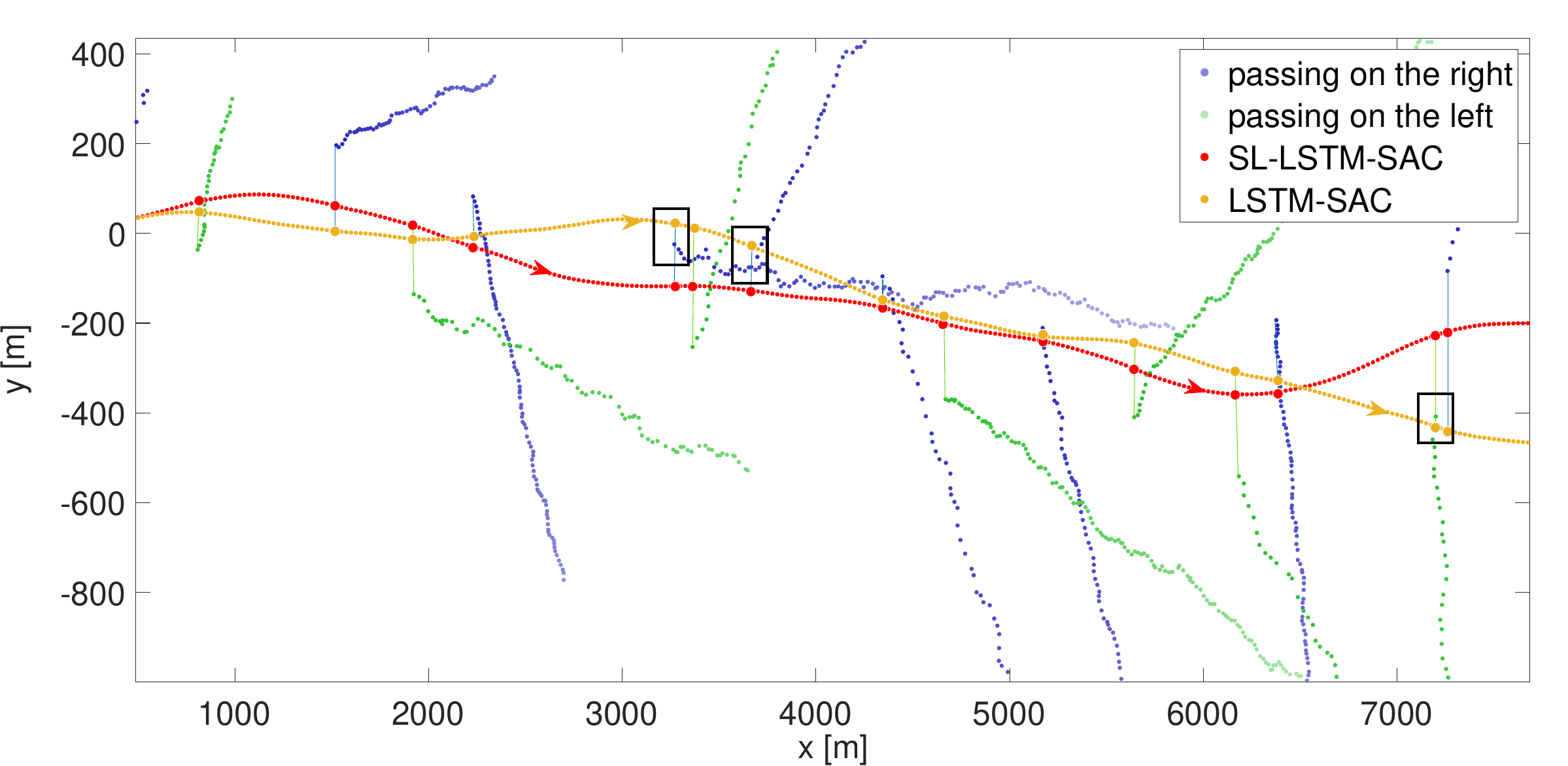}
	\caption[Agent trajectories with stochastic obstacle movement]{Agent trajectories in the obstacle avoidance environment with stochastic obstacle movement, black squares indicating a violation of passing rules.}
	\label{fig:env_example_AR1}
\end{figure}

\begin{figure}[ht!]
	\centering
	\includegraphics[width=1\linewidth]{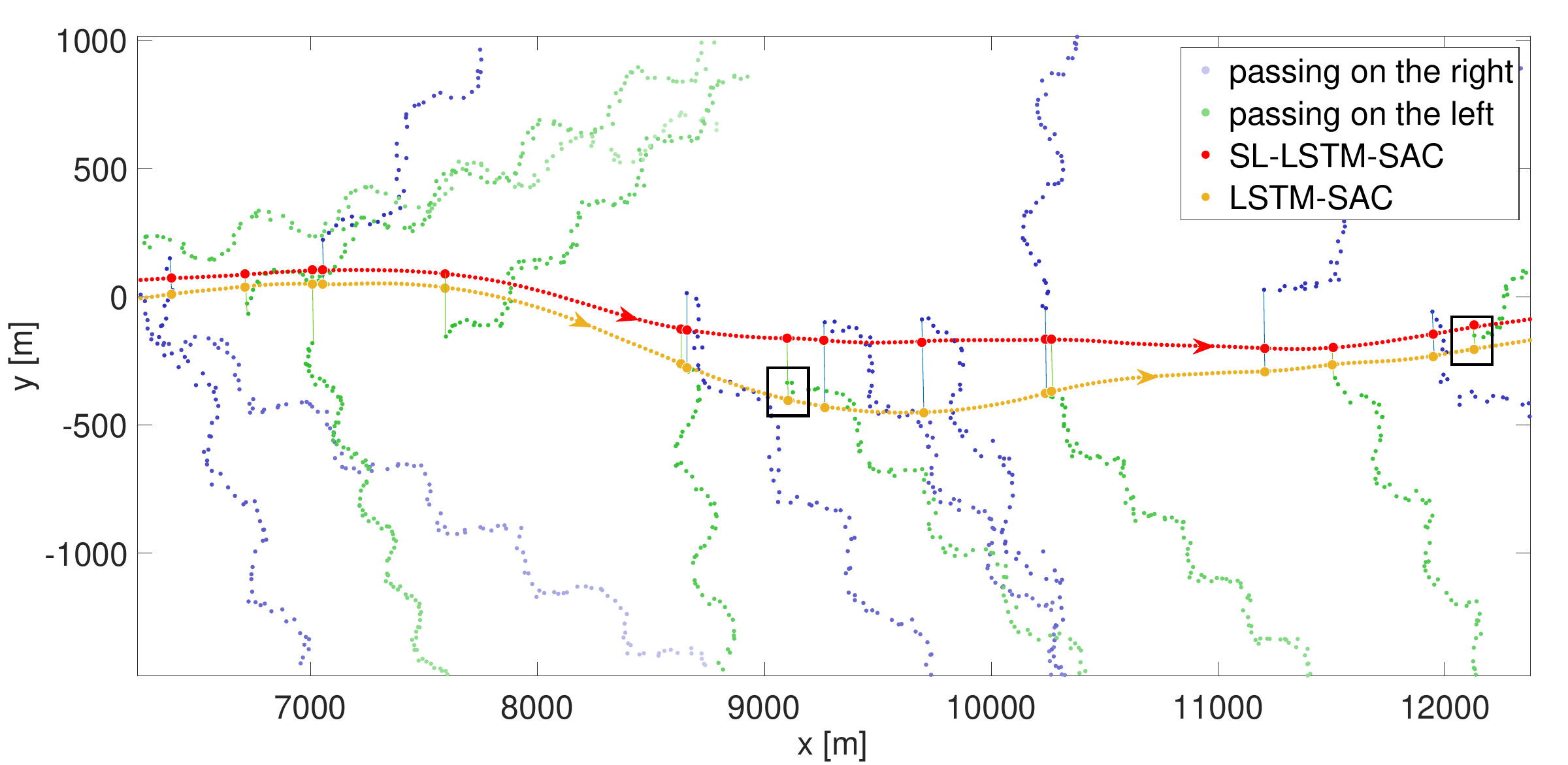}
	\caption[Agent trajectories with periodic obstacle movement]{Agent trajectories in the obstacle avoidance environment with periodic obstacle movement, black squares indicating a violation of passing rules.}
	\label{fig:env_example_sinus}
\end{figure}

We visualize an exemplary episode for stochastic and periodic obstacle movements in Figures \ref{fig:env_example_AR1} and \ref{fig:env_example_sinus}, respectively, to give further intuition about how the trained agents perform in the obstacle avoidance environment. \rdd{For illustration purposes, we show the trajectories of the LSTM-SAC with and without the SL module.} The agent trajectories are depicted in red and yellow, and obstacle trajectories in blue and green, respectively, according to the imposed passing rules. The violations of the passing rules are marked with a black square. \rdd{We observe that the agent equipped with the additional CR estimates (prefix 'SL') consistently selects better trajectories, avoiding violations of passing rules. In contrast, the baseline agent struggles to accurately predict the trajectories of some obstacles, leading to rule violations.}

\rdd{For instance, in Figure \ref{fig:env_example_AR1}, shortly after $x = \unit[3000]{m}$, the SL-LSTM-SAC agent successfully navigates past an upcoming obstacle on the right, whereas the baseline agent's lateral position is too high, resulting in a rule violation. Later in the episode, after $x = \unit[7000]{m}$, the SL-LSTM-SAC agent correctly passes an obstacle on the left and another on the right, while the baseline agent violates the rules by passing both obstacles on the right side. Similar findings are evident in Figure \ref{fig:env_example_sinus}. For example, at a longitudinal position slightly beyond $x = \unit[9000]{m}$, the baseline agent incorrectly crosses an obstacle on the right side, violating the passing rules, while the SL-LSTM-SAC agent correctly passes on the left-hand side. These examples illustrate the superiority of the SL-LSTM-SAC agent in adhering to passing rules and choosing adequate trajectories.}

\section{\rdd{Generalization study: Maritime traffic}}\label{sec:gen_study}
\subsection{\rdd{Trajectory data}}
\rdd{We demonstrate the generalizability of our two-step proposal to DOA by investigating the maritime traffic domain. In particular, we design an environment where the agent controls a vessel and needs to resolve head-on situations with high \rdd{CR}. Head-on scenarios are maritime encounter situations where two vessels approach each other on reciprocal or nearly reciprocal courses, posing a high risk of collision if neither vessel takes evasive action \citep{COLREGs1972}. The obstacle vessel follows a real-world trajectory obtained from the Automatic Identification System (AIS, \citealt{yang2019big}), whose data we thankfully received from the European Maritime Safety Agency. The AIS is a tracking system used to identify and localize vessels, containing information such as positions, speeds, headings, timestamps, destinations, and ship types, among other details. Passenger ships and cargo ships above certain sizes must carry an AIS transponder and emit data to the system \citep{solas1974}.}

\begin{figure}[htp]
    \centering
    \includegraphics[width=\textwidth]{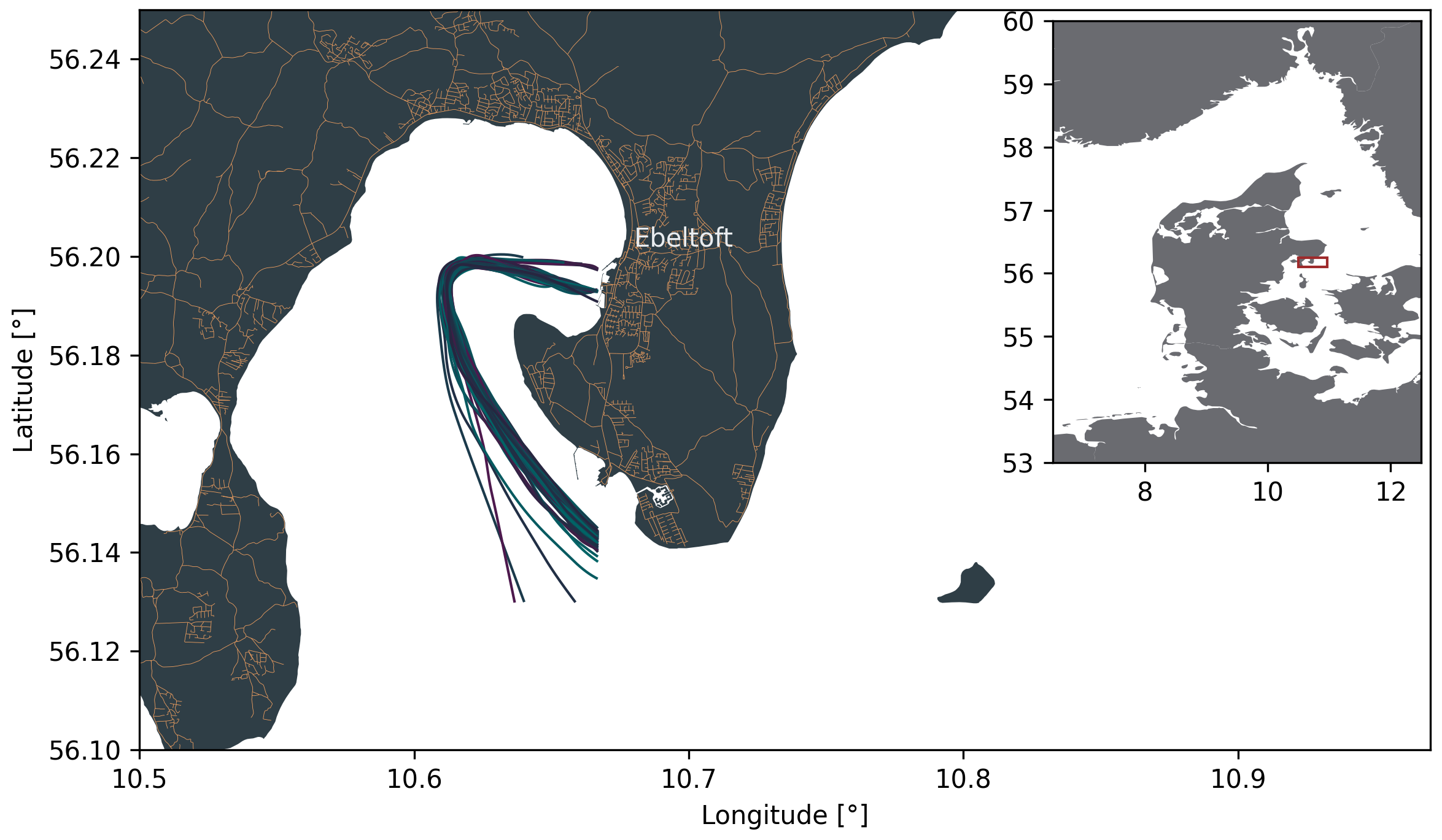}
    \caption[AIS trajectories close to Ebeltoft, Denmark]{\rdd{The real-world AIS trajectories in the maritime area close to Ebeltoft, Denmark.}}
    \label{fig:Ebeltoft_trajs}
\end{figure}

\rdd{Acknowledging that AIS records may contain redundant or incomplete information \citep{last2014comprehensive}, we use the package of \cite{paulig2024open} to employ filtering techniques to eliminate duplicate records and entries with missing information.~As a use case, we filter vessel trajectories from the year 2021 in a specific area close to the city of Ebeltoft in Denmark. In total, 137 different trajectories have been extracted, and subsequent waypoints in a trajectory are interpolated based on cubic splines to be $\Delta t = \unit[5]{s}$ apart. As visualized in Figure \ref{fig:Ebeltoft_trajs}, the area features vessel trajectories characterized by highly non-linear behavior. Vessels need to navigate a curve to reach the open sea from the Ebeltoft harbor, making it a suitable candidate to demonstrate our two-step approach to DOA under non-linear obstacle movements.}

\subsection{\rdd{Supervised learning}}\label{subsec:SL_vessels}
\rdd{The supervised learning module adheres to the trajectory prediction and \rdd{CR} estimation framework outlined in Section \rrd{\ref{sec:Implementation}}. However, accounting for the increased complexity of the real-world trajectories, we empirically determined that two adjustments became necessary to maintain a high prediction quality. First, we increase the network size. In Figure \ref{fig:Architecture}, the trajectory prediction network comprises an LSTM layer with 64 hidden units, followed by two linear layers with 64 neurons each, using ReLU and linear activation functions, respectively. For this case study, we expand the LSTM layer to 256 hidden units and employ three linear layers with 256 neurons each. Additionally, the first two linear layers now use leaky ReLU instead of a conventional ReLU activation function \citep{maas2013rectifier}.}

\rdd{Second, the trajectory prediction network described in Section \rrd{\ref{sec:trajectory_prediction}} fed the LSTM unit with the first differences of the last 10 obstacle positions \((\vec{p}_{t-9}, \dots, \vec{p}_t)^\top\). The output of the LSTM was then passed through the linear layers to estimate the next obstacle position in \(\mathbb{R}^2\). In contrast, in this section, we recursively process the relative differences of the last 75 obstacle positions \((\vec{p}_{t-74}, \dots, \vec{p}_t)^\top\) using the LSTM. Considering a time step consists of $\unit[5]{s}$, 75 observations equate to information about the past $\unit[6]{min}$ $\unit[25]{s}$, a realistic time horizon for vessel traffic.} 

\rdd{In addition, we concatenate the LSTM's output with the absolute position \(\vec{p}_t\). This process can be understood as a form of 'anchoring'. As illustrated in Figure \ref{fig:Ebeltoft_trajs}, absolute position information is necessary to determine whether a curve should be forecasted, despite the general stability of predicting relative positions. Thus, we opt for this hybrid approach to leverage the strengths of both methods. Additionally, to save computation time due to the increased network size, we directly forecast the next 10 relative obstacle positions. Consequently, the new trajectory prediction module outputs in \(\mathbb{R}^{10 \cdot 2} = \mathbb{R}^{20}\).}

\begin{figure}[ht]
  \centering
  \begin{subfigure}[b]{0.49\textwidth}
    \centering
    \includegraphics[width=\textwidth]{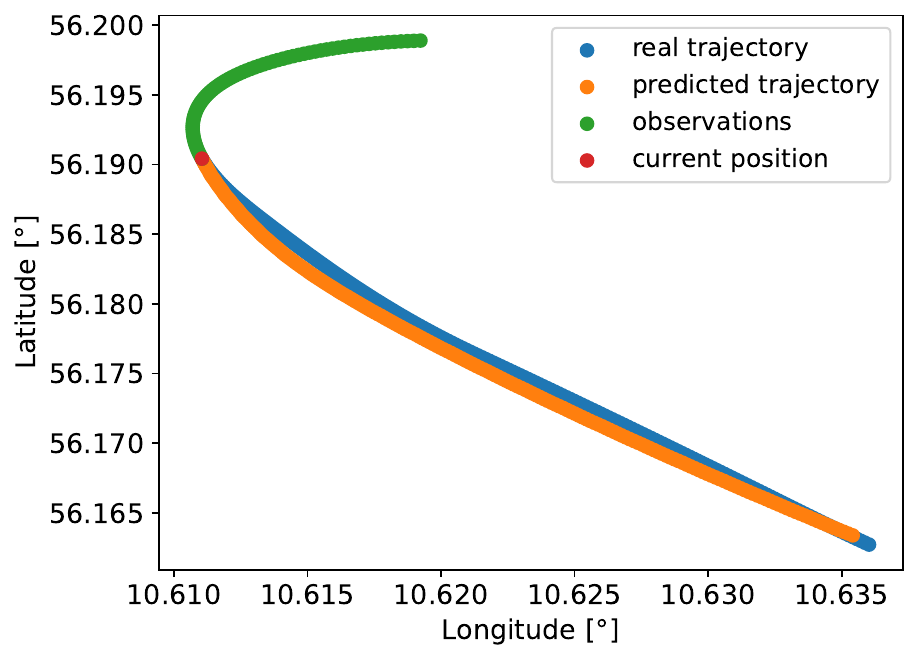}
  \end{subfigure}
  \hfill
  \begin{subfigure}[b]{0.49\textwidth}
    \centering
    \includegraphics[width=\textwidth]{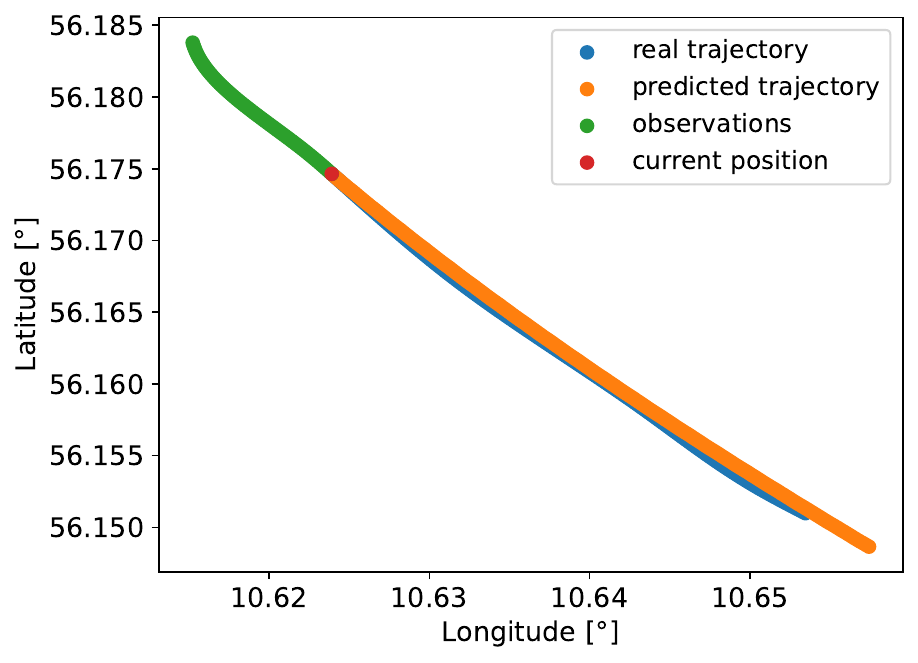}
  \end{subfigure}
  \vfill
  \centering
  \begin{subfigure}[b]{0.49\textwidth}
    \centering
    \includegraphics[width=\textwidth]{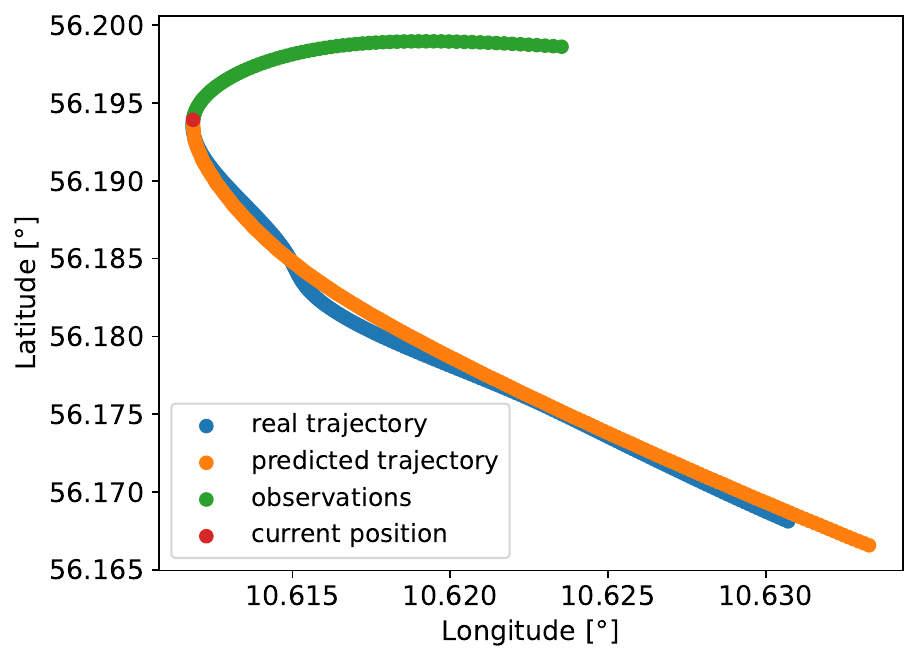}
  \end{subfigure}
  \hfill
  \begin{subfigure}[b]{0.49\textwidth}
    \centering
    \includegraphics[width=\textwidth]{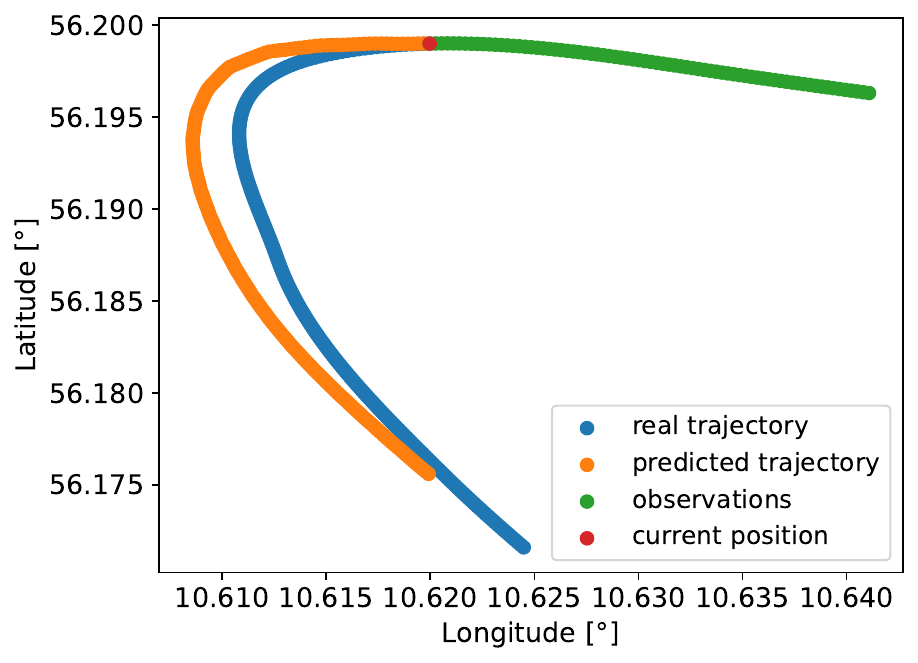}
  \end{subfigure}
  \caption[Example trajectory estimates using supervised learning]{\rdd{Examples of the trajectory estimates using the supervised learning module.}}
  \label{fig:traj_AIS_single_estimates}
\end{figure}

\rdd{During training, we randomly sample segments from the trajectories and add Gaussian noise to the positions to reduce overfitting. Visualizations of the resulting trajectory estimates after training are shown in Figure \ref{fig:traj_AIS_single_estimates}. Overall, the predictions are accurate and capture the non-linear behavior well. However, the necessity of the anchoring mechanism is evident in the lower right plot of Figure \ref{fig:traj_AIS_single_estimates}. The observations in green, used to make predictions in orange, show no signs of a curve, making it difficult for the neural network to anticipate when the curve will appear. Without the anchoring, forecasting the curve would be almost impossible.}

\begin{figure}[htp!]
    \centering
    \includegraphics[width=0.65\textwidth]{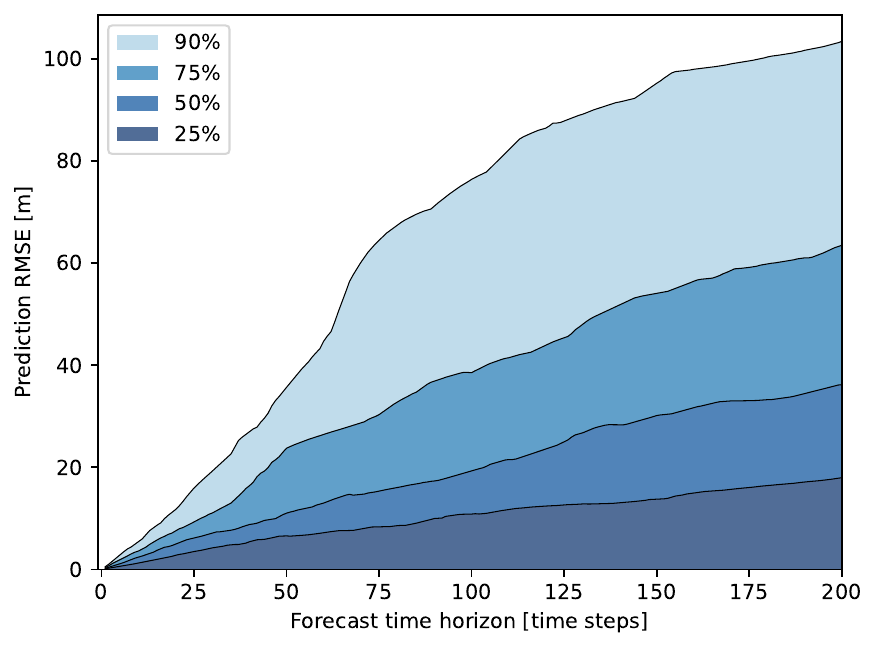}
    \caption[RMSE of vessel trajectory predictions]{\rdd{RMSE between predicted and actual trajectory for different quantiles.}}
    \label{fig:quantile_forecast_vessel}
\end{figure}

\rdd{Similar to Figure \ref{fig:prediction_rmse}, we include a quantitative evaluation based on the RMSE over time in Figure \ref{fig:quantile_forecast_vessel}. In particular, we randomly sample 100 trajectory segments, forecast the next 200 time steps, and compare the predictions with the true trajectory points. Note that 200 time steps equate approximately $\unit[16.67]{min}$. Following Figure \ref{fig:quantile_forecast_vessel}, we have a median RMSE of approximately $\unit[60]{m}$ over the time horizon.}

\subsection{\rdd{Reinforcement learning environment}}
\begin{figure}[ht]
  \centering
  \begin{subfigure}[b]{0.49\textwidth}
    \centering
    \includegraphics[width=\textwidth]{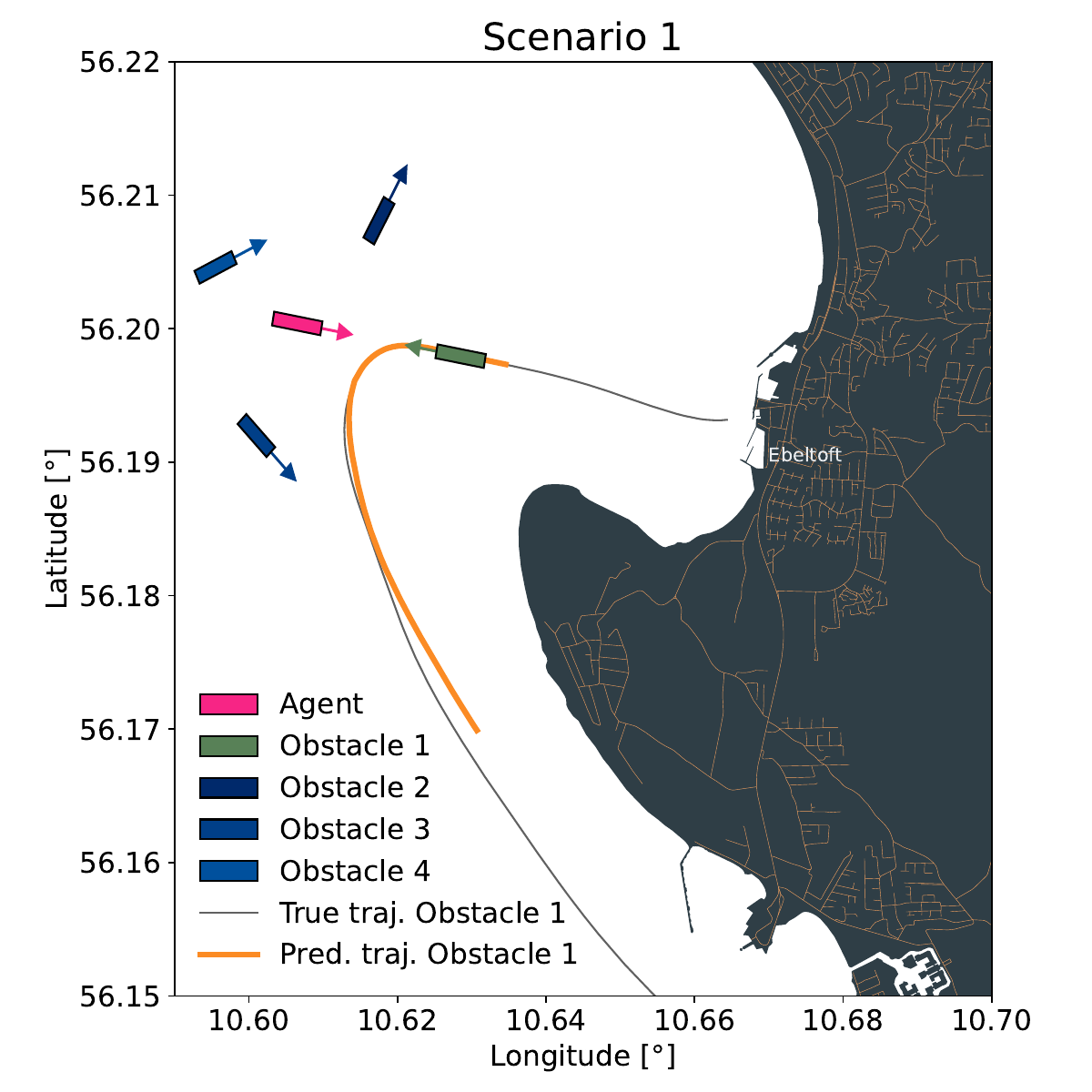}
  \end{subfigure}
  \hfill
  \begin{subfigure}[b]{0.49\textwidth}
    \centering
    \includegraphics[width=\textwidth]{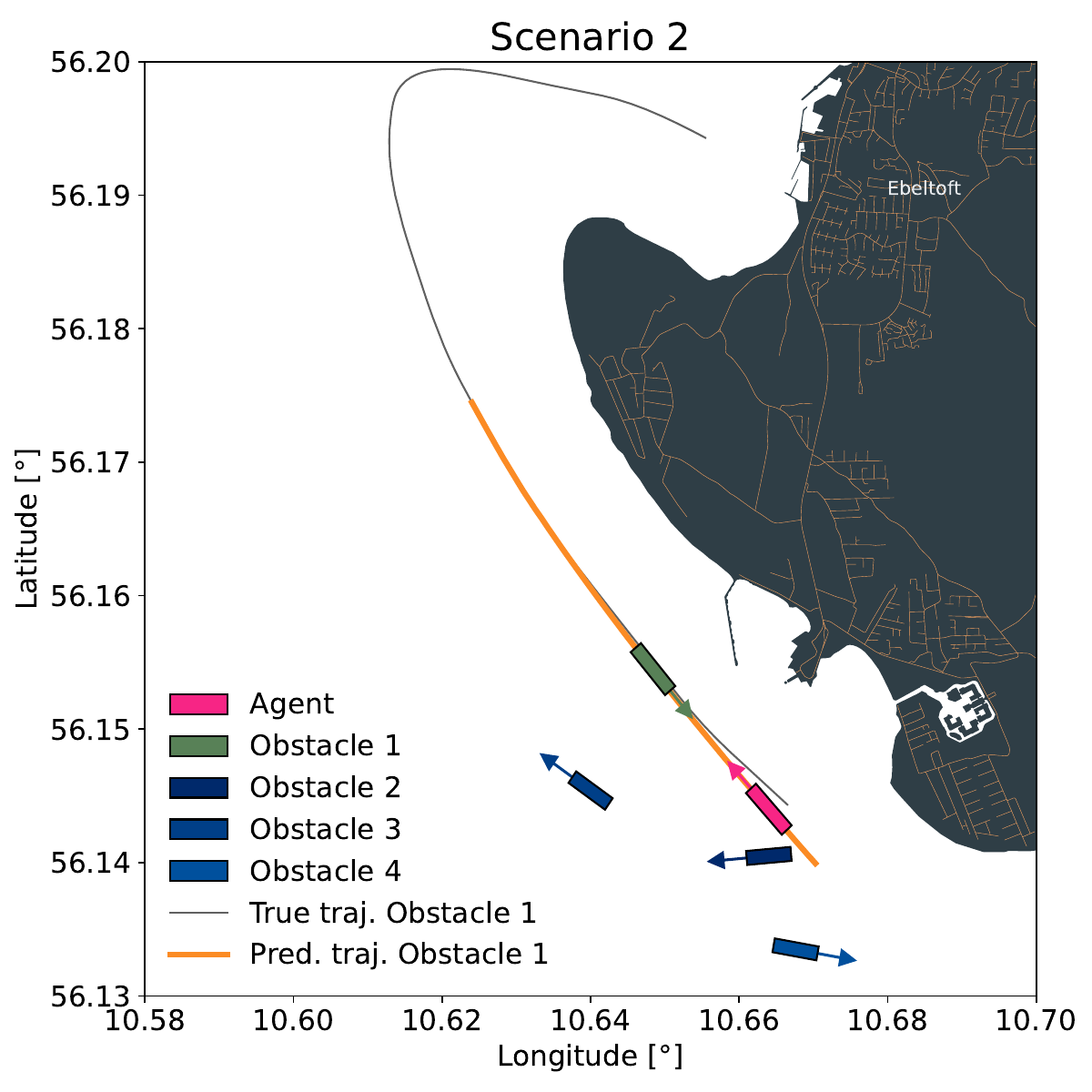}
  \end{subfigure}
  \caption[Maritime traffic environment]{\rdd{The two possible scenarios in the maritime traffic environment. The grey solid line is the true trajectory of Obstacle 1, while the predicted trajectory using the SL module is shown in orange. Obstacle 1 performs a portside curve in the left plot while it travels almost straight in the right plot.}}
  \label{fig:AIS_env_situations}
\end{figure}

\rdd{The maritime RL environment requires the agent to resolve head-on situations with an obstacle vessel, as visualized in Figure \ref{fig:AIS_env_situations}. At the beginning of each episode, we randomly sample an obstacle trajectory, corresponding to Obstacle 1 in the figure. We then determine, with equal probability, whether Obstacle 1 is initialized before or after the curve, depicted in the left or right plot in Figure \ref{fig:AIS_env_situations}, respectively. In both scenarios, the agent is penalized if it avoids Obstacle 1 via a portside (left) turn since such a maneuver does not comply with maritime traffic rules for head-on situations \citep{COLREGs1972}. Consequently, the agent can develop the following strategy: If Obstacle 1 is set to perform a curve, the agent can maintain its course; otherwise, it should execute an avoidance maneuver to the starboard (right) side.}

\rdd{In addition, we randomly generate three other obstacles in each episode, which move linearly and are included to increase the complexity of the problem. The agent has a speed of approximately \unit[8]{m/s} and is initialized with a time-to-collision of \unit[4]{min} to the head-on vessel. Furthermore, we add Gaussian noise to the initial positions of the agent and obstacle vessels to ensure the diversity of the episodes. As in Section \ref{sec:results}, we compare RL agents with the \rdd{CR} estimates from the SL module with baseline agents without those estimates.}

\rdd{The observation vector of an agent at time $t$ is defined as follows:}
\rdd{\begin{equation}\label{eq:o_t_maritime}
	o_{t} = \begin{pmatrix}
		\frac{U_{t,\rm agent}}{U_{\rm scale}},
        \frac{y^{e}_{t, \rm agent}}{y^{e}_{\rm scale}},
		\frac{\Vec{d}_{t,i}-d_{\rm coll}}{d_{\rm scale}},
        \frac{\Vec{\alpha}_{t,i}}{\pi},
        \frac{\Vec{\psi}_{t,i}-\psi_{t, \rm agent}}{\pi},
        \frac{\vec{U}_{t,i}}{U_{\rm scale}},
        \frac{\hat{d}^{\rm CPA}_{t,i} - d_{\rm coll}}{d^{\rm CPA}_{\rm norm}},
        \frac{\hat{t}^{\rm CPA}_{t,i}}{t^{\rm CPA}_{\rm norm}}
        \end{pmatrix}^\top,
\end{equation}
where $U_{t,\rm agent}$ and $\psi_{t, \rm agent}$ are the speed and heading angle of the agent. The cross-track error of the agent is denoted $y^{e}_{t, \rm agent}$ and constitutes the lateral offset to the linear path from the agent's starting position during an episode. This metric informs the agent whether it performs a starboard or portside avoidance maneuver. Furthermore, $\vec{d}_{t,i}$ is the Euclidean distance between the agent and obstacle $i$, $\Vec{\psi}_{t,i}$ and $\vec{U}_{t,i}$ denote the heading angle and speed of obstacle $i$, and $\vec{\alpha}_{t,i}$ is the relative bearing of obstacle $i$ from the perspective of the agent. For definitions of these metrics, we refer to the textbook of \cite{fossen2021handbook}.}

\rdd{As in Section \rrd{\ref{subsec:RL_env}}, $\hat{d}^{\rm CPA}_{t,i}$ and $\hat{t}^{\rm CPA}_{t,i}$ denote the estimates of the $d^{\rm CPA}$ and $t^{\rm CPA}$ with obstacle $i$ from the SL module. The normalizing constants are set to $U_{\rm scale}= \unit[5]{m/s}$, $y^{e}_{\rm scale} = \unit[200]{m}$, $d_{\rm coll} = \unit[300]{m}$, $d_{\rm scale} = \unit[3000]{m}$, $d^{\rm CPA}_{\rm norm} = \unit[100]{m}$, and $t^{\rm CPA}_{\rm norm} = \unit[100]{s}$. Considering that there are four obstacles in the environment, the length of $o_{t}$ is $2 + 4 \cdot 6 = 26$. We sort $o_{t}$ with respect to the ascending estimated $d^{\rm CPA}$. For the baselines without the CR estimates, we sort $o_{t}$ with respect to the ascending Euclidean distance. We emphasize that only the \rdd{CR} estimates for Obstacle 1 in Figure \ref{fig:AIS_env_situations} are computed via the SL module. Since the other obstacles move linearly, using the analytic formulas of \cite{lenart1983collision} suffices.}

\rdd{The action of an agent at time $t$, $a_{t}$, modifies the agent's heading:
\begin{equation*}
    \psi_{t+1, \rm agent} = \psi_{t, \rm agent} + a_{t} \cdot a_{\rm scale},
\end{equation*}
where the value $a_{\rm scale} = 2.4^\circ$ is carefully chosen to ensure the agent's maneuverability is limited enough to create a difficult situation while ensuring the problem remains solvable. Furthermore, we allow the agent to select an action only every four time steps to ensure the maneuverability is not unrealistically high.}

\rdd{Lastly, the reward function is computed via:
\begin{equation*}
    r_{t+1} = - \mathbf{1}\{y^{e}_{t} \geq y^{e}_{\rm scale}\} - \sum_{i=1}^{4}\mathbf{1}\{d_{i,t} \leq d_{\rm coll}\},
\end{equation*}
with the indicator function $\mathbf{1}\{x\}$ being one if $x$ is true and zero otherwise. The first term is the penalty for performing a portside maneuver, while the sum term relates to collision avoidance. An episode ends when 60 time steps, equating to \unit[5]{min}, have passed.}

\subsection{\rdd{Results}}\label{subsubsec:results_AIS}
\subsubsection{\rdd{Setup}}
\rdd{We run the LSTM-TD3 and LSTM-SAC algorithms for both configurations, with and without the estimates from the SL module. The hyperparameters are identical to those used in Section \ref{sec:results} and are listed in Table \ref{tab:hyperparams_RL}. Additionally, we set the history length of the RL algorithms to \(l=4\) as it showed the best performance in Figure \ref{fig:h_ablate}. The return development during training is displayed in Figure \ref{fig:AIS_results}, where we average the returns of 15 independent runs and include 95\% point-wise confidence intervals.}

\rdd{Furthermore, we investigate the robustness of the final policies of the different agents in the presence of positional noise. Specifically, at each time step, we disturb the sensed north and east positions of each involved vessel with independent realizations of a univariate Gaussian variable with zero mean and standard deviations \(\sigma \in \{0, 5, 10, 20\}\) meters to reflect real-world sensor noise \citep{liu2016unmanned}. The state features in (\ref{eq:o_t_maritime}) were subsequently computed using the noisy positional observations. For each algorithm, noise level, and scenario in the environment, we run 100 randomly initialized test episodes and measure whether a collision occurred. The collision rates are presented in Figure \ref{fig:noise_analysis}, including a confidence interval based on the 100 observations.}

\subsubsection{\rdd{Analysis}}
\begin{figure}[ht!]
    \centering
    \includegraphics[width=0.6\linewidth]{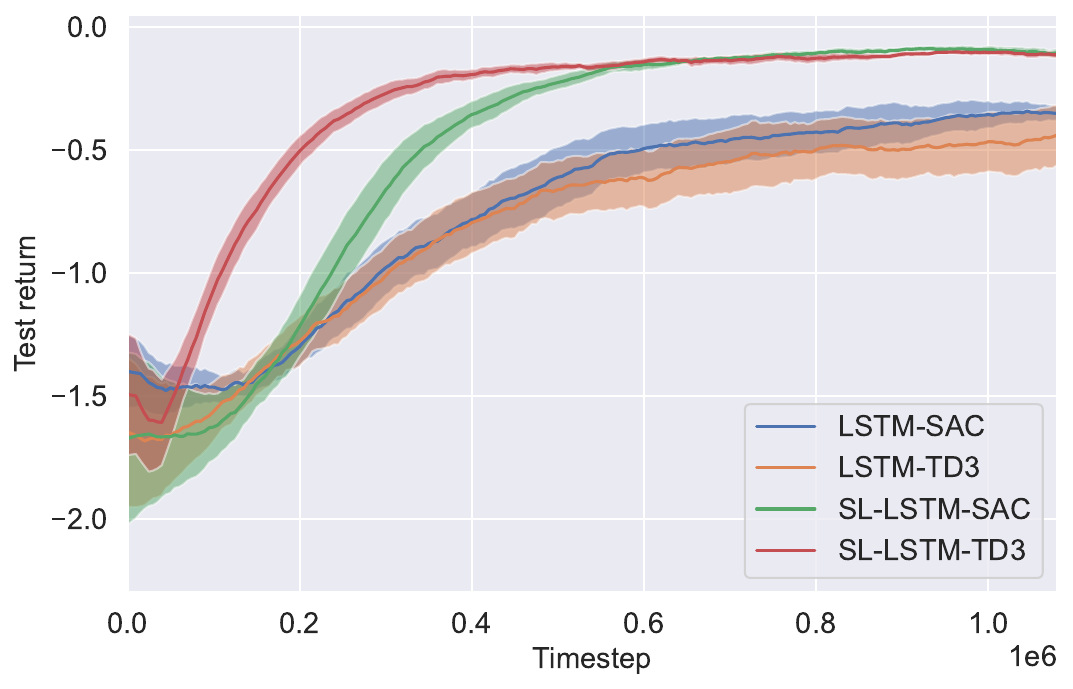}
    \caption[Performance comparison for maritime traffic environment]{\rdd{Performance comparison of agents that use the estimates for $d^{\rm CPA}$ and $t^{\rm CPA}$ (prefix 'SL') with baseline agents without the SL subroutine for the maritime traffic environment. The y-axis displays the test return during an episode.}}
    \label{fig:AIS_results}
\end{figure}

\begin{figure}[ht!]
    \centering
    \includegraphics[width=0.9\linewidth]{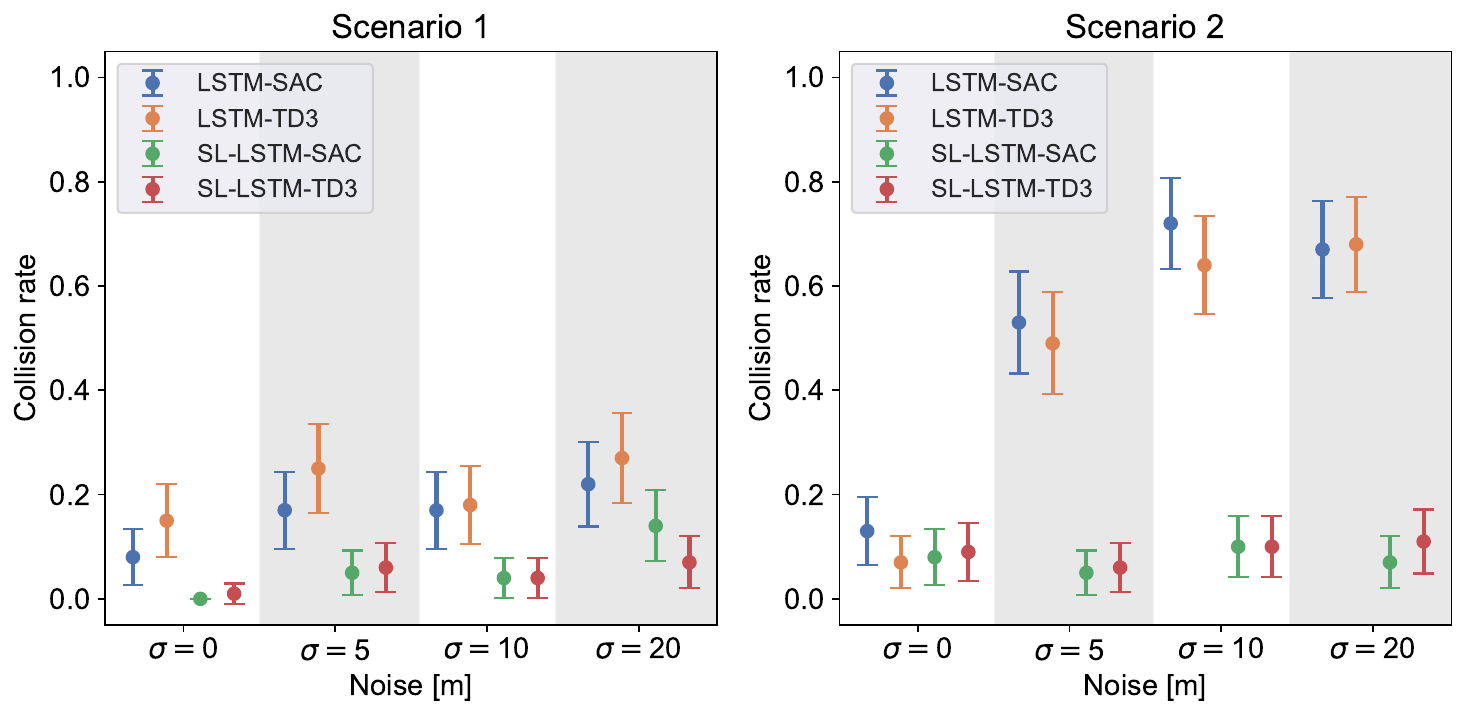}
    \caption[Robustness analysis]{\rdd{Robustness analysis of the agents with respect to positional noise.}}
    \label{fig:noise_analysis}
\end{figure}

\rdd{Similar to the findings in Section \ref{sec:results}, Figure \ref{fig:AIS_results} demonstrates that the configurations with the SL prefix significantly outperform their conventional counterparts in terms of final return performance. Additionally, these configurations exhibit faster convergence and reduced variance in test returns. Moreover, Figure \ref{fig:noise_analysis} highlights the increased robustness when leveraging the SL estimates. Both approaches, with and without SL, generally learn the correct strategies of maintaining the course in Scenario 1 and turning starboard in Scenario 2. However, even in the case $\sigma = \unit[0]{m}$, which corresponds to the noise-free training scenarios, the DRL agents without the SL estimates more frequently misinterpret Scenario 1 and produce a collision. Moreover, the SL-based approaches show remarkable resilience to increased positional noise. In contrast, the performance of the conventional approaches degrades significantly, resulting in much higher collision rates, especially in Scenario 2.}

\section{\rdd{Discussion and outlook}}\label{sec:discussion}
\rdd{The results of the investigations in Sections \ref{sec:results} and \ref{sec:gen_study} display the performance potential of our proposed two-step approach to DOA. In particular, DRL is especially promising for real-time DOA tasks due to its adaptability and capability of handling a wide range of environments and obstacle configurations without explicit programming for each scenario. In addition, the fast inference time of neural networks makes our approach highly practical for real-time DOA situations.}

\rdd{However, we acknowledge the limitations of our work that can be addressed in future research. In particular, we focus on non-cooperative obstacle behavior, where the obstacles themselves do not perform active collision avoidance maneuvers. Our two-step architecture could handle such scenarios if historical data on obstacle behavior in specific encounter situations is available, and future research may validate the proposal in such cases. Furthermore, the static nature of historical data does not account for potential changes in obstacle behavior over time due to evolving dynamics or new traffic patterns. Implementing a mechanism to periodically update the supervised learning model with new data and trends would ensure that the system remains effective and relevant in changing environments. This continuous learning approach would allow the model to adapt to new obstacle behaviors as they emerge, maintaining high levels of safety and efficiency in obstacle avoidance.}

\rdd{In addition, we focus on non-communicative situations where the obstacles do not explicitly inform the agent about their intentions, necessitating the prediction of their future trajectories. If such information channels were available, as assumed in \cite{akdaug2022collaborative}, the decision-making of the DRL agent can be further improved. Nevertheless, a non-communicative system, like ours, is still essential to handle cases when the communication channel fails \citep{chen2017decentralized}. Lastly, our proposal provides point estimates of the $d^{\rm CPA}$ and $t^{\rm CPA}$ risk metrics. While this procedure strongly improves the performance of the respective DRL agent, we acknowledge that the \rdd{CR} estimates critically depend on the uncertainty connected to the predicted obstacle trajectories. To address this, it might be beneficial to derive confidence bounds for the \rdd{CR} estimates by considering sampling-based approaches \citep{lee2017desire} or imposing distributional assumptions \citep{chai2020multipath} for the trajectory prediction. This addition might allow for better-informed decision-making by quantifying the reliability of the risk assessments and accounting for potential variations in obstacle behavior.}

\section{Conclusion}\label{sec:conclusion}
Future sustainable transportation systems can benefit from autonomous vehicles in terms of safety and energy efficiency, particularly in non-lane-based domains such as air or maritime traffic. Such real-world autonomous systems necessitate robust local path planning capabilities, including the critical task of avoiding collisions with dynamic obstacles. This work proposes a two-step architecture that leverages recent advances from supervised learning and reinforcement learning. The first step captures non-linear movements of obstacles and estimates \rdd{CR} metrics in a supervised fashion, while the second step uses these estimates to enrich the observation space of an RL agent. 

Our approach has demonstrated a significant enhancement in the collision avoidance capabilities of the respective autonomous agent, thereby constituting a valuable contribution to the realm of autonomous vehicles. Importantly, our learning architecture exhibits versatility by not being confined to a specific set of obstacle dynamics. \rdd{We validated our approach in a generic environment, independent of a particular traffic domain, and a maritime traffic environment using real-world vessel trajectory data.} In our upcoming research, we aim to validate this architecture using a \rdd{wider} range of real-world obstacle behavior data and implement it across different traffic domains. Moreover, \rdd{we will continue to improve the proposed methodology via the routes outlined in Section \ref{sec:discussion}, including the consideration of cooperative obstacle behavior and uncertainty quantification of the \rdd{CR} estimates.}

\section*{Acknowledgement}
 This work was partially funded by BAW - Bundesanstalt für Wasserbau (Mikrosimulation des Schiffsverkehrs auf dem Niederrhein, award number 181.7.03\_0004\_02\_2023). The authors are grateful to the Center for Information Services and High Performance Computing at TU Dresden for providing its facilities for high throughput calculations. \rdd{Furthermore, the authors thank the European Maritime Safety Agency for providing the AIS data used for the generalization study of maritime traffic.}
\clearpage
\bibliographystyle{elsarticle-harv} 
\bibliography{bib}

\begin{thebibliography}{109}
\expandafter\ifx\csname natexlab\endcsname\relax\def\natexlab#1{#1}\fi
\providecommand{\url}[1]{\texttt{#1}}
\providecommand{\href}[2]{#2}
\providecommand{\path}[1]{#1}
\providecommand{\DOIprefix}{doi:}
\providecommand{\ArXivprefix}{arXiv:}
\providecommand{\URLprefix}{URL: }
\providecommand{\Pubmedprefix}{pmid:}
\providecommand{\doi}[1]{\href{http://dx.doi.org/#1}{\path{#1}}}
\providecommand{\Pubmed}[1]{\href{pmid:#1}{\path{#1}}}
\providecommand{\bibinfo}[2]{#2}
\ifx\xfnm\relax \def\xfnm[#1]{\unskip,\space#1}\fi
\bibitem[{Akda{\u{g}} et~al.(2022)Akda{\u{g}}, Fossen and Johansen}]{akdaug2022collaborative}
\bibinfo{author}{Akda{\u{g}}, M.}, \bibinfo{author}{Fossen, T.I.}, \bibinfo{author}{Johansen, T.A.}, \bibinfo{year}{2022}.
\newblock \bibinfo{title}{Collaborative collision avoidance for autonomous ships using informed scenario-based model predictive control}.
\newblock \bibinfo{journal}{IFAC-PapersOnLine} \bibinfo{volume}{55}, \bibinfo{pages}{249--256}.
\bibitem[{Badue et~al.(2021)Badue, Guidolini, Carneiro, Azevedo, Cardoso, Forechi, Jesus, Berriel, Paixao, Mutz et~al.}]{badue2021self}
\bibinfo{author}{Badue, C.}, \bibinfo{author}{Guidolini, R.}, \bibinfo{author}{Carneiro, R.V.}, \bibinfo{author}{Azevedo, P.}, \bibinfo{author}{Cardoso, V.B.}, \bibinfo{author}{Forechi, A.}, \bibinfo{author}{Jesus, L.}, \bibinfo{author}{Berriel, R.}, \bibinfo{author}{Paixao, T.M.}, \bibinfo{author}{Mutz, F.}, et~al., \bibinfo{year}{2021}.
\newblock \bibinfo{title}{Self-driving cars: A survey}.
\newblock \bibinfo{journal}{Expert Systems with Applications} \bibinfo{volume}{165}, \bibinfo{pages}{113816}.
\bibitem[{Bellemare et~al.(2020)Bellemare, Candido, Castro, Gong, Machado, Moitra, Ponda and Wang}]{bellemare2020autonomous}
\bibinfo{author}{Bellemare, M.G.}, \bibinfo{author}{Candido, S.}, \bibinfo{author}{Castro, P.S.}, \bibinfo{author}{Gong, J.}, \bibinfo{author}{Machado, M.C.}, \bibinfo{author}{Moitra, S.}, \bibinfo{author}{Ponda, S.S.}, \bibinfo{author}{Wang, Z.}, \bibinfo{year}{2020}.
\newblock \bibinfo{title}{Autonomous navigation of stratospheric balloons using reinforcement learning}.
\newblock \bibinfo{journal}{Nature} \bibinfo{volume}{588}, \bibinfo{pages}{77--82}.
\bibitem[{Borenstein et~al.(1991)Borenstein, Koren et~al.}]{borenstein1991vector}
\bibinfo{author}{Borenstein, J.}, \bibinfo{author}{Koren, Y.}, et~al., \bibinfo{year}{1991}.
\newblock \bibinfo{title}{The vector field histogram-fast obstacle avoidance for mobile robots}.
\newblock \bibinfo{journal}{IEEE Transactions on Robotics and Automation} \bibinfo{volume}{7}, \bibinfo{pages}{278--288}.
\bibitem[{Brittain and Wei(2022)}]{brittain2022scalable}
\bibinfo{author}{Brittain, M.}, \bibinfo{author}{Wei, P.}, \bibinfo{year}{2022}.
\newblock \bibinfo{title}{Scalable autonomous separation assurance with heterogeneous multi-agent reinforcement learning}.
\newblock \bibinfo{journal}{IEEE Transactions on Automation Science and Engineering} \bibinfo{volume}{19}, \bibinfo{pages}{2837--2848}.
\bibitem[{Chai et~al.(2020)Chai, Sapp, Bansal and Anguelov}]{chai2020multipath}
\bibinfo{author}{Chai, Y.}, \bibinfo{author}{Sapp, B.}, \bibinfo{author}{Bansal, M.}, \bibinfo{author}{Anguelov, D.}, \bibinfo{year}{2020}.
\newblock \bibinfo{title}{Multipath: Multiple probabilistic anchor trajectory hypotheses for behavior prediction}, in: \bibinfo{booktitle}{Conference on Robot Learning}, \bibinfo{organization}{PMLR}. pp. \bibinfo{pages}{86--99}.
\bibitem[{Chen et~al.(2019)Chen, Huang, Mou and Van~Gelder}]{chen2019probabilistic}
\bibinfo{author}{Chen, P.}, \bibinfo{author}{Huang, Y.}, \bibinfo{author}{Mou, J.}, \bibinfo{author}{Van~Gelder, P.}, \bibinfo{year}{2019}.
\newblock \bibinfo{title}{Probabilistic risk analysis for ship-ship collision: State-of-the-art}.
\newblock \bibinfo{journal}{Safety science} \bibinfo{volume}{117}, \bibinfo{pages}{108--122}.
\bibitem[{Chen et~al.(2023)Chen, Yao, McAuley, Zhou and Wang}]{chen2023deep}
\bibinfo{author}{Chen, X.}, \bibinfo{author}{Yao, L.}, \bibinfo{author}{McAuley, J.}, \bibinfo{author}{Zhou, G.}, \bibinfo{author}{Wang, X.}, \bibinfo{year}{2023}.
\newblock \bibinfo{title}{Deep reinforcement learning in recommender systems: A survey and new perspectives}.
\newblock \bibinfo{journal}{Knowledge-Based Systems} \bibinfo{volume}{264}, \bibinfo{pages}{110335}.
\bibitem[{Chen et~al.(2017)Chen, Liu, Everett and How}]{chen2017decentralized}
\bibinfo{author}{Chen, Y.F.}, \bibinfo{author}{Liu, M.}, \bibinfo{author}{Everett, M.}, \bibinfo{author}{How, J.P.}, \bibinfo{year}{2017}.
\newblock \bibinfo{title}{Decentralized non-communicating multiagent collision avoidance with deep reinforcement learning}, in: \bibinfo{booktitle}{International Conference on Robotics and Automation}, \bibinfo{organization}{IEEE}. pp. \bibinfo{pages}{285--292}.
\bibitem[{Chun et~al.(2021)Chun, Roh, Lee, Ha and Yu}]{chun2021deep}
\bibinfo{author}{Chun, D.H.}, \bibinfo{author}{Roh, M.I.}, \bibinfo{author}{Lee, H.W.}, \bibinfo{author}{Ha, J.}, \bibinfo{author}{Yu, D.}, \bibinfo{year}{2021}.
\newblock \bibinfo{title}{Deep reinforcement learning-based collision avoidance for an autonomous ship}.
\newblock \bibinfo{journal}{Ocean Engineering} \bibinfo{volume}{234}, \bibinfo{pages}{109216}.
\bibitem[{Debnath and Chin(2010)}]{debnath2010navigational}
\bibinfo{author}{Debnath, A.K.}, \bibinfo{author}{Chin, H.C.}, \bibinfo{year}{2010}.
\newblock \bibinfo{title}{Navigational traffic conflict technique: a proactive approach to quantitative measurement of collision risks in port waters}.
\newblock \bibinfo{journal}{The Journal of Navigation} \bibinfo{volume}{63}, \bibinfo{pages}{137--152}.
\bibitem[{Dhillon and Verma(2020)}]{dhillon2020convolutional}
\bibinfo{author}{Dhillon, A.}, \bibinfo{author}{Verma, G.K.}, \bibinfo{year}{2020}.
\newblock \bibinfo{title}{Convolutional neural network: a review of models, methodologies and applications to object detection}.
\newblock \bibinfo{journal}{Progress in Artificial Intelligence} \bibinfo{volume}{9}, \bibinfo{pages}{85--112}.
\bibitem[{Dorigo and Gambardella(1997)}]{dorigo1997ant}
\bibinfo{author}{Dorigo, M.}, \bibinfo{author}{Gambardella, L.M.}, \bibinfo{year}{1997}.
\newblock \bibinfo{title}{Ant colonies for the travelling salesman problem}.
\newblock \bibinfo{journal}{Biosystems} \bibinfo{volume}{43}, \bibinfo{pages}{73--81}.
\bibitem[{Duguleana and Mogan(2016)}]{duguleana2016neural}
\bibinfo{author}{Duguleana, M.}, \bibinfo{author}{Mogan, G.}, \bibinfo{year}{2016}.
\newblock \bibinfo{title}{Neural networks based reinforcement learning for mobile robots obstacle avoidance}.
\newblock \bibinfo{journal}{Expert Systems with Applications} \bibinfo{volume}{62}, \bibinfo{pages}{104--115}.
\bibitem[{Everett et~al.(2018)Everett, Chen and How}]{everett2018motion}
\bibinfo{author}{Everett, M.}, \bibinfo{author}{Chen, Y.F.}, \bibinfo{author}{How, J.P.}, \bibinfo{year}{2018}.
\newblock \bibinfo{title}{Motion planning among dynamic, decision-making agents with deep reinforcement learning}, in: \bibinfo{booktitle}{International Conference on Intelligent Robots and Systems}, \bibinfo{organization}{IEEE}. pp. \bibinfo{pages}{3052--3059}.
\bibitem[{Everett et~al.(2021)Everett, Chen and How}]{everett2021collision}
\bibinfo{author}{Everett, M.}, \bibinfo{author}{Chen, Y.F.}, \bibinfo{author}{How, J.P.}, \bibinfo{year}{2021}.
\newblock \bibinfo{title}{Collision avoidance in pedestrian-rich environments with deep reinforcement learning}.
\newblock \bibinfo{journal}{IEEE Access} \bibinfo{volume}{9}, \bibinfo{pages}{10357--10377}.
\bibitem[{Falanga et~al.(2020)Falanga, Kleber and Scaramuzza}]{falanga2020dynamic}
\bibinfo{author}{Falanga, D.}, \bibinfo{author}{Kleber, K.}, \bibinfo{author}{Scaramuzza, D.}, \bibinfo{year}{2020}.
\newblock \bibinfo{title}{Dynamic obstacle avoidance for quadrotors with event cameras}.
\newblock \bibinfo{journal}{Science Robotics} \bibinfo{volume}{5}, \bibinfo{pages}{eaaz9712}.
\bibitem[{Fawzi et~al.(2022)Fawzi, Balog, Huang, Hubert, Romera-Paredes, Barekatain, Novikov, R~Ruiz, Schrittwieser, Swirszcz et~al.}]{fawzi2022discovering}
\bibinfo{author}{Fawzi, A.}, \bibinfo{author}{Balog, M.}, \bibinfo{author}{Huang, A.}, \bibinfo{author}{Hubert, T.}, \bibinfo{author}{Romera-Paredes, B.}, \bibinfo{author}{Barekatain, M.}, \bibinfo{author}{Novikov, A.}, \bibinfo{author}{R~Ruiz, F.J.}, \bibinfo{author}{Schrittwieser, J.}, \bibinfo{author}{Swirszcz, G.}, et~al., \bibinfo{year}{2022}.
\newblock \bibinfo{title}{Discovering faster matrix multiplication algorithms with reinforcement learning}.
\newblock \bibinfo{journal}{Nature} \bibinfo{volume}{610}, \bibinfo{pages}{47--53}.
\bibitem[{Feng et~al.(2023)Feng, Sun, Yan, Zhu, Zou, Shen and Liu}]{feng2023dense}
\bibinfo{author}{Feng, S.}, \bibinfo{author}{Sun, H.}, \bibinfo{author}{Yan, X.}, \bibinfo{author}{Zhu, H.}, \bibinfo{author}{Zou, Z.}, \bibinfo{author}{Shen, S.}, \bibinfo{author}{Liu, H.X.}, \bibinfo{year}{2023}.
\newblock \bibinfo{title}{Dense reinforcement learning for safety validation of autonomous vehicles}.
\newblock \bibinfo{journal}{Nature} \bibinfo{volume}{615}, \bibinfo{pages}{620--627}.
\bibitem[{Fiorini and Shiller(1998)}]{fiorini1998motion}
\bibinfo{author}{Fiorini, P.}, \bibinfo{author}{Shiller, Z.}, \bibinfo{year}{1998}.
\newblock \bibinfo{title}{Motion planning in dynamic environments using velocity obstacles}.
\newblock \bibinfo{journal}{The International Journal of Robotics Research} \bibinfo{volume}{17}, \bibinfo{pages}{760--772}.
\bibitem[{Fossen(2021)}]{fossen2021handbook}
\bibinfo{author}{Fossen, T.I.}, \bibinfo{year}{2021}.
\newblock \bibinfo{title}{Handbook of Marine Craft Hydrodynamics and Motion Control, 2nd Edition}.
\newblock \bibinfo{publisher}{John Wiley \& Sons}.
\bibitem[{Fox et~al.(1997)Fox, Burgard and Thrun}]{fox1997dynamic}
\bibinfo{author}{Fox, D.}, \bibinfo{author}{Burgard, W.}, \bibinfo{author}{Thrun, S.}, \bibinfo{year}{1997}.
\newblock \bibinfo{title}{The dynamic window approach to collision avoidance}.
\newblock \bibinfo{journal}{IEEE Robotics \& Automation Magazine} \bibinfo{volume}{4}, \bibinfo{pages}{23--33}.
\bibitem[{Fujii and Tanaka(1971)}]{fujii1971traffic}
\bibinfo{author}{Fujii, Y.}, \bibinfo{author}{Tanaka, K.}, \bibinfo{year}{1971}.
\newblock \bibinfo{title}{Traffic capacity}.
\newblock \bibinfo{journal}{The Journal of Navigation} \bibinfo{volume}{24}, \bibinfo{pages}{543--552}.
\bibitem[{Fujimoto et~al.(2018)Fujimoto, Hoof and Meger}]{fujimoto2018addressing}
\bibinfo{author}{Fujimoto, S.}, \bibinfo{author}{Hoof, H.}, \bibinfo{author}{Meger, D.}, \bibinfo{year}{2018}.
\newblock \bibinfo{title}{Addressing function approximation error in actor-critic methods}, in: \bibinfo{booktitle}{International Conference on Machine Learning}, \bibinfo{organization}{PMLR}. pp. \bibinfo{pages}{1587--1596}.
\bibitem[{Haarnoja et~al.(2018a)Haarnoja, Zhou, Abbeel and Levine}]{haarnoja2018soft}
\bibinfo{author}{Haarnoja, T.}, \bibinfo{author}{Zhou, A.}, \bibinfo{author}{Abbeel, P.}, \bibinfo{author}{Levine, S.}, \bibinfo{year}{2018}a.
\newblock \bibinfo{title}{Soft actor-critic: Off-policy maximum entropy deep reinforcement learning with a stochastic actor}, in: \bibinfo{booktitle}{International Conference on Machine Learning}, \bibinfo{organization}{PMLR}. pp. \bibinfo{pages}{1861--1870}.
\bibitem[{Haarnoja et~al.(2018b)Haarnoja, Zhou, Hartikainen, Tucker, Ha, Tan, Kumar, Zhu, Gupta, Abbeel et~al.}]{haarnoja2018soft2}
\bibinfo{author}{Haarnoja, T.}, \bibinfo{author}{Zhou, A.}, \bibinfo{author}{Hartikainen, K.}, \bibinfo{author}{Tucker, G.}, \bibinfo{author}{Ha, S.}, \bibinfo{author}{Tan, J.}, \bibinfo{author}{Kumar, V.}, \bibinfo{author}{Zhu, H.}, \bibinfo{author}{Gupta, A.}, \bibinfo{author}{Abbeel, P.}, et~al., \bibinfo{year}{2018}b.
\newblock \bibinfo{title}{Soft actor-critic algorithms and applications}.
\newblock \bibinfo{journal}{arXiv preprint arXiv:1812.05905} .
\bibitem[{Hansen et~al.(2013)Hansen, Jensen, Lehn-Schi{\o}ler, Melchild, Rasmussen and Ennemark}]{hansen2013empirical}
\bibinfo{author}{Hansen, M.G.}, \bibinfo{author}{Jensen, T.K.}, \bibinfo{author}{Lehn-Schi{\o}ler, T.}, \bibinfo{author}{Melchild, K.}, \bibinfo{author}{Rasmussen, F.M.}, \bibinfo{author}{Ennemark, F.}, \bibinfo{year}{2013}.
\newblock \bibinfo{title}{Empirical ship domain based on ais data}.
\newblock \bibinfo{journal}{The Journal of Navigation} \bibinfo{volume}{66}, \bibinfo{pages}{931--940}.
\bibitem[{Hart and Okhrin(2024)}]{hart2022enhanced}
\bibinfo{author}{Hart, F.}, \bibinfo{author}{Okhrin, O.}, \bibinfo{year}{2024}.
\newblock \bibinfo{title}{Enhanced method for reinforcement learning based dynamic obstacle avoidance by assessment of collision risk}.
\newblock \bibinfo{journal}{Neurocomputing} \bibinfo{volume}{568}, \bibinfo{pages}{127097}.
\bibitem[{Heess et~al.(2015)Heess, Hunt, Lillicrap and Silver}]{heess2015memory}
\bibinfo{author}{Heess, N.}, \bibinfo{author}{Hunt, J.J.}, \bibinfo{author}{Lillicrap, T.P.}, \bibinfo{author}{Silver, D.}, \bibinfo{year}{2015}.
\newblock \bibinfo{title}{Memory-based control with recurrent neural networks}.
\newblock \bibinfo{journal}{arXiv preprint arXiv:1512.04455} .
\bibitem[{Heiberg et~al.(2022)Heiberg, Larsen, Meyer, Rasheed, San and Varagnolo}]{heiberg2022risk}
\bibinfo{author}{Heiberg, A.}, \bibinfo{author}{Larsen, T.N.}, \bibinfo{author}{Meyer, E.}, \bibinfo{author}{Rasheed, A.}, \bibinfo{author}{San, O.}, \bibinfo{author}{Varagnolo, D.}, \bibinfo{year}{2022}.
\newblock \bibinfo{title}{Risk-based implementation of {COLREGs} for autonomous surface vehicles using deep reinforcement learning}.
\newblock \bibinfo{journal}{Neural Networks} \bibinfo{volume}{152}, \bibinfo{pages}{17--33}.
\bibitem[{Hochreiter and Schmidhuber(1997)}]{hochreiter1997long}
\bibinfo{author}{Hochreiter, S.}, \bibinfo{author}{Schmidhuber, J.}, \bibinfo{year}{1997}.
\newblock \bibinfo{title}{Long short-term memory}.
\newblock \bibinfo{journal}{Neural computation} \bibinfo{volume}{9}, \bibinfo{pages}{1735--1780}.
\bibitem[{Hu et~al.(2023)Hu, Tian, Zhao and Shen}]{hu2023path}
\bibinfo{author}{Hu, S.}, \bibinfo{author}{Tian, S.}, \bibinfo{author}{Zhao, J.}, \bibinfo{author}{Shen, R.}, \bibinfo{year}{2023}.
\newblock \bibinfo{title}{Path planning of an unmanned surface vessel based on the improved a-star and dynamic window method}.
\newblock \bibinfo{journal}{Journal of Marine Science and Engineering} \bibinfo{volume}{11}, \bibinfo{pages}{1060}.
\bibitem[{Huang et~al.(2020)Huang, Chen, Chen, Negenborn and Van~Gelder}]{huang2020ship}
\bibinfo{author}{Huang, Y.}, \bibinfo{author}{Chen, L.}, \bibinfo{author}{Chen, P.}, \bibinfo{author}{Negenborn, R.R.}, \bibinfo{author}{Van~Gelder, P.}, \bibinfo{year}{2020}.
\newblock \bibinfo{title}{Ship collision avoidance methods: State-of-the-art}.
\newblock \bibinfo{journal}{Safety Science} \bibinfo{volume}{121}, \bibinfo{pages}{451--473}.
\bibitem[{Hwang and Seah(2008)}]{hwang2008intent}
\bibinfo{author}{Hwang, I.}, \bibinfo{author}{Seah, C.E.}, \bibinfo{year}{2008}.
\newblock \bibinfo{title}{Intent-based probabilistic conflict detection for the next generation air transportation system}.
\newblock \bibinfo{journal}{Proceedings of the IEEE} \bibinfo{volume}{96}, \bibinfo{pages}{2040--2059}.
\bibitem[{{International Maritime Organization}(1972)}]{COLREGs1972}
\bibinfo{author}{{International Maritime Organization}}, \bibinfo{year}{1972}.
\newblock \bibinfo{title}{{COLREG}: Convention on the International Regulations for Preventing Collisions at Sea}.
\bibitem[{{International Maritime Organization}(1974)}]{solas1974}
\bibinfo{author}{{International Maritime Organization}}, \bibinfo{year}{1974}.
\newblock \bibinfo{title}{{SOLAS}: International Convention for the Safety of Life at Sea}.
\bibitem[{Isufaj et~al.(2022)Isufaj, Omeri and Piera}]{isufaj2022multi}
\bibinfo{author}{Isufaj, R.}, \bibinfo{author}{Omeri, M.}, \bibinfo{author}{Piera, M.A.}, \bibinfo{year}{2022}.
\newblock \bibinfo{title}{Multi-{UAV} conflict resolution with graph convolutional reinforcement learning}.
\newblock \bibinfo{journal}{Applied Sciences} \bibinfo{volume}{12}, \bibinfo{pages}{610}.
\bibitem[{Jiang et~al.(2022)Jiang, An, Zhang, Wang and Wang}]{jiang2022human}
\bibinfo{author}{Jiang, L.}, \bibinfo{author}{An, L.}, \bibinfo{author}{Zhang, X.}, \bibinfo{author}{Wang, C.}, \bibinfo{author}{Wang, X.}, \bibinfo{year}{2022}.
\newblock \bibinfo{title}{A human-like collision avoidance method for autonomous ship with attention-based deep reinforcement learning}.
\newblock \bibinfo{journal}{Ocean Engineering} \bibinfo{volume}{264}, \bibinfo{pages}{112378}.
\bibitem[{Kaelbling et~al.(1998)Kaelbling, Littman and Cassandra}]{kaelbling1998planning}
\bibinfo{author}{Kaelbling, L.P.}, \bibinfo{author}{Littman, M.L.}, \bibinfo{author}{Cassandra, A.R.}, \bibinfo{year}{1998}.
\newblock \bibinfo{title}{Planning and acting in partially observable stochastic domains}.
\newblock \bibinfo{journal}{Artificial Intelligence} \bibinfo{volume}{101}, \bibinfo{pages}{99--134}.
\bibitem[{Kandepu et~al.(2008)Kandepu, Foss and Imsland}]{kandepu2008applying}
\bibinfo{author}{Kandepu, R.}, \bibinfo{author}{Foss, B.}, \bibinfo{author}{Imsland, L.}, \bibinfo{year}{2008}.
\newblock \bibinfo{title}{Applying the unscented kalman filter for nonlinear state estimation}.
\newblock \bibinfo{journal}{Journal of Process Control} \bibinfo{volume}{18}, \bibinfo{pages}{753--768}.
\bibitem[{Kennedy and Eberhart(1995)}]{kennedy1995particle}
\bibinfo{author}{Kennedy, J.}, \bibinfo{author}{Eberhart, R.}, \bibinfo{year}{1995}.
\newblock \bibinfo{title}{Particle swarm optimization}, in: \bibinfo{booktitle}{International Conference on Neural Networks}, \bibinfo{organization}{IEEE}. pp. \bibinfo{pages}{1942--1948}.
\bibitem[{Khatib(1986)}]{khatib1986real}
\bibinfo{author}{Khatib, O.}, \bibinfo{year}{1986}.
\newblock \bibinfo{title}{Real-time obstacle avoidance for manipulators and mobile robots}.
\newblock \bibinfo{journal}{The International Journal of Robotics Research} \bibinfo{volume}{5}, \bibinfo{pages}{90--98}.
\bibitem[{Kingma and Ba(2014)}]{kingma2014adam}
\bibinfo{author}{Kingma, D.P.}, \bibinfo{author}{Ba, J.}, \bibinfo{year}{2014}.
\newblock \bibinfo{title}{Adam: A method for stochastic optimization}.
\newblock \bibinfo{journal}{arXiv preprint arXiv:1412.6980} .
\bibitem[{Kroese et~al.(2006)Kroese, Porotsky and Rubinstein}]{kroese2006cross}
\bibinfo{author}{Kroese, D.P.}, \bibinfo{author}{Porotsky, S.}, \bibinfo{author}{Rubinstein, R.Y.}, \bibinfo{year}{2006}.
\newblock \bibinfo{title}{The cross-entropy method for continuous multi-extremal optimization}.
\newblock \bibinfo{journal}{Methodology and Computing in Applied Probability} \bibinfo{volume}{8}, \bibinfo{pages}{383--407}.
\bibitem[{Kuchar and Yang(2000)}]{kuchar2000review}
\bibinfo{author}{Kuchar, J.K.}, \bibinfo{author}{Yang, L.C.}, \bibinfo{year}{2000}.
\newblock \bibinfo{title}{A review of conflict detection and resolution modeling methods}.
\newblock \bibinfo{journal}{IEEE Transactions On Intelligent Transportation Systems} \bibinfo{volume}{1}, \bibinfo{pages}{179--189}.
\bibitem[{Kuwata et~al.(2013)Kuwata, Wolf, Zarzhitsky and Huntsberger}]{kuwata2013safe}
\bibinfo{author}{Kuwata, Y.}, \bibinfo{author}{Wolf, M.T.}, \bibinfo{author}{Zarzhitsky, D.}, \bibinfo{author}{Huntsberger, T.L.}, \bibinfo{year}{2013}.
\newblock \bibinfo{title}{Safe maritime autonomous navigation with {COLREGS}, using velocity obstacles}.
\newblock \bibinfo{journal}{IEEE Journal of Oceanic Engineering} \bibinfo{volume}{39}, \bibinfo{pages}{110--119}.
\bibitem[{Last et~al.(2014)Last, Bahlke, Hering-Bertram and Linsen}]{last2014comprehensive}
\bibinfo{author}{Last, P.}, \bibinfo{author}{Bahlke, C.}, \bibinfo{author}{Hering-Bertram, M.}, \bibinfo{author}{Linsen, L.}, \bibinfo{year}{2014}.
\newblock \bibinfo{title}{Comprehensive analysis of automatic identification system ({AIS}) data in regard to vessel movement prediction}.
\newblock \bibinfo{journal}{The Journal of Navigation} \bibinfo{volume}{67}, \bibinfo{pages}{791--809}.
\bibitem[{Latombe(2012)}]{latombe2012robot}
\bibinfo{author}{Latombe, J.C.}, \bibinfo{year}{2012}.
\newblock \bibinfo{title}{Robot motion planning}. volume \bibinfo{volume}{124}.
\newblock \bibinfo{publisher}{Springer Science \& Business Media}.
\bibitem[{LaValle et~al.(2001)LaValle, Kuffner, Donald et~al.}]{lavalle2001rapidly}
\bibinfo{author}{LaValle, S.M.}, \bibinfo{author}{Kuffner, J.J.}, \bibinfo{author}{Donald, B.}, et~al., \bibinfo{year}{2001}.
\newblock \bibinfo{title}{Rapidly-exploring random trees: Progress and prospects}.
\newblock \bibinfo{journal}{Algorithmic and computational robotics: new directions} \bibinfo{volume}{5}, \bibinfo{pages}{293--308}.
\bibitem[{LeCun et~al.(2015)LeCun, Bengio and Hinton}]{lecun2015deep}
\bibinfo{author}{LeCun, Y.}, \bibinfo{author}{Bengio, Y.}, \bibinfo{author}{Hinton, G.}, \bibinfo{year}{2015}.
\newblock \bibinfo{title}{Deep learning}.
\newblock \bibinfo{journal}{Nature} \bibinfo{volume}{521}, \bibinfo{pages}{436--444}.
\bibitem[{Lee et~al.(2019)Lee, Zhu, Srinivasan, Shah, Savarese, Fei-Fei, Garg and Bohg}]{lee2019making}
\bibinfo{author}{Lee, M.A.}, \bibinfo{author}{Zhu, Y.}, \bibinfo{author}{Srinivasan, K.}, \bibinfo{author}{Shah, P.}, \bibinfo{author}{Savarese, S.}, \bibinfo{author}{Fei-Fei, L.}, \bibinfo{author}{Garg, A.}, \bibinfo{author}{Bohg, J.}, \bibinfo{year}{2019}.
\newblock \bibinfo{title}{Making sense of vision and touch: Self-supervised learning of multimodal representations for contact-rich tasks}, in: \bibinfo{booktitle}{International Conference on Robotics and Automation}, \bibinfo{organization}{IEEE}. pp. \bibinfo{pages}{8943--8950}.
\bibitem[{Lee et~al.(2017)Lee, Choi, Vernaza, Choy, Torr and Chandraker}]{lee2017desire}
\bibinfo{author}{Lee, N.}, \bibinfo{author}{Choi, W.}, \bibinfo{author}{Vernaza, P.}, \bibinfo{author}{Choy, C.B.}, \bibinfo{author}{Torr, P.H.}, \bibinfo{author}{Chandraker, M.}, \bibinfo{year}{2017}.
\newblock \bibinfo{title}{Desire: Distant future prediction in dynamic scenes with interacting agents}, in: \bibinfo{booktitle}{Proceedings of the IEEE conference on computer vision and pattern recognition}, pp. \bibinfo{pages}{336--345}.
\bibitem[{Lenart(1983)}]{lenart1983collision}
\bibinfo{author}{Lenart, A.S.}, \bibinfo{year}{1983}.
\newblock \bibinfo{title}{Collision threat parameters for a new radar display and plot technique}.
\newblock \bibinfo{journal}{The Journal of Navigation} \bibinfo{volume}{36}, \bibinfo{pages}{404--410}.
\bibitem[{Li(2019)}]{li2019reinforcement}
\bibinfo{author}{Li, Y.}, \bibinfo{year}{2019}.
\newblock \bibinfo{title}{Reinforcement learning applications}.
\newblock \bibinfo{journal}{arXiv preprint arXiv:1908.06973} .
\bibitem[{Liu et~al.(2023)Liu, Qiu, Yang, Wang, Xiang, Wang and Xu}]{liu2023colregs}
\bibinfo{author}{Liu, W.}, \bibinfo{author}{Qiu, K.}, \bibinfo{author}{Yang, X.}, \bibinfo{author}{Wang, R.}, \bibinfo{author}{Xiang, Z.}, \bibinfo{author}{Wang, Y.}, \bibinfo{author}{Xu, W.}, \bibinfo{year}{2023}.
\newblock \bibinfo{title}{{COLREGS}-based collision avoidance algorithm for unmanned surface vehicles using modified artificial potential fields}.
\newblock \bibinfo{journal}{Physical Communication} \bibinfo{volume}{57}, \bibinfo{pages}{101980}.
\bibitem[{Liu et~al.(2016)Liu, Zhang, Yu and Yuan}]{liu2016unmanned}
\bibinfo{author}{Liu, Z.}, \bibinfo{author}{Zhang, Y.}, \bibinfo{author}{Yu, X.}, \bibinfo{author}{Yuan, C.}, \bibinfo{year}{2016}.
\newblock \bibinfo{title}{Unmanned surface vehicles: An overview of developments and challenges}.
\newblock \bibinfo{journal}{Annual Reviews in Control} \bibinfo{volume}{41}, \bibinfo{pages}{71--93}.
\bibitem[{Lyu and Yin(2019)}]{lyu2019colregs}
\bibinfo{author}{Lyu, H.}, \bibinfo{author}{Yin, Y.}, \bibinfo{year}{2019}.
\newblock \bibinfo{title}{{COLREGS}-constrained real-time path planning for autonomous ships using modified artificial potential fields}.
\newblock \bibinfo{journal}{The Journal of Navigation} \bibinfo{volume}{72}, \bibinfo{pages}{588--608}.
\bibitem[{Maas et~al.(2013)Maas, Hannun, Ng et~al.}]{maas2013rectifier}
\bibinfo{author}{Maas, A.L.}, \bibinfo{author}{Hannun, A.Y.}, \bibinfo{author}{Ng, A.Y.}, et~al., \bibinfo{year}{2013}.
\newblock \bibinfo{title}{Rectifier nonlinearities improve neural network acoustic models}, in: \bibinfo{booktitle}{International Conference on Machine Learning}, \bibinfo{organization}{Atlanta, GA}. p.~\bibinfo{pages}{3}.
\bibitem[{Marin-Plaza et~al.(2018)Marin-Plaza, Hussein, Martin and Escalera}]{marin2018global}
\bibinfo{author}{Marin-Plaza, P.}, \bibinfo{author}{Hussein, A.}, \bibinfo{author}{Martin, D.}, \bibinfo{author}{Escalera, A.d.l.}, \bibinfo{year}{2018}.
\newblock \bibinfo{title}{Global and local path planning study in a {ROS}-based research platform for autonomous vehicles}.
\newblock \bibinfo{journal}{Journal of Advanced Transportation} \bibinfo{volume}{2018}.
\bibitem[{Meng et~al.(2021)Meng, Gorbet and Kuli{\'c}}]{meng2021memory}
\bibinfo{author}{Meng, L.}, \bibinfo{author}{Gorbet, R.}, \bibinfo{author}{Kuli{\'c}, D.}, \bibinfo{year}{2021}.
\newblock \bibinfo{title}{Memory-based deep reinforcement learning for pomdps}, in: \bibinfo{booktitle}{International Conference on Intelligent Robots and Systems}, \bibinfo{organization}{IEEE}. pp. \bibinfo{pages}{5619--5626}.
\bibitem[{Meyer et~al.(2020)Meyer, Robinson, Rasheed and San}]{meyer2020taming}
\bibinfo{author}{Meyer, E.}, \bibinfo{author}{Robinson, H.}, \bibinfo{author}{Rasheed, A.}, \bibinfo{author}{San, O.}, \bibinfo{year}{2020}.
\newblock \bibinfo{title}{Taming an autonomous surface vehicle for path following and collision avoidance using deep reinforcement learning}.
\newblock \bibinfo{journal}{IEEE Access} \bibinfo{volume}{8}, \bibinfo{pages}{41466--41481}.
\bibitem[{Mitici and Blom(2018)}]{mitici2018mathematical}
\bibinfo{author}{Mitici, M.}, \bibinfo{author}{Blom, H.A.}, \bibinfo{year}{2018}.
\newblock \bibinfo{title}{Mathematical models for air traffic conflict and collision probability estimation}.
\newblock \bibinfo{journal}{IEEE Transactions on Intelligent Transportation Systems} \bibinfo{volume}{20}, \bibinfo{pages}{1052--1068}.
\bibitem[{Mnih et~al.(2015)Mnih, Kavukcuoglu, Silver, Rusu, Veness, Bellemare, Graves, Riedmiller, Fidjeland, Ostrovski et~al.}]{mnih2015human}
\bibinfo{author}{Mnih, V.}, \bibinfo{author}{Kavukcuoglu, K.}, \bibinfo{author}{Silver, D.}, \bibinfo{author}{Rusu, A.A.}, \bibinfo{author}{Veness, J.}, \bibinfo{author}{Bellemare, M.G.}, \bibinfo{author}{Graves, A.}, \bibinfo{author}{Riedmiller, M.}, \bibinfo{author}{Fidjeland, A.K.}, \bibinfo{author}{Ostrovski, G.}, et~al., \bibinfo{year}{2015}.
\newblock \bibinfo{title}{Human-level control through deep reinforcement learning}.
\newblock \bibinfo{journal}{Nature} \bibinfo{volume}{518}, \bibinfo{pages}{529--533}.
\bibitem[{Mou et~al.(2010)Mou, Van Der~Tak and Ligteringen}]{mou2010study}
\bibinfo{author}{Mou, J.M.}, \bibinfo{author}{Van Der~Tak, C.}, \bibinfo{author}{Ligteringen, H.}, \bibinfo{year}{2010}.
\newblock \bibinfo{title}{Study on collision avoidance in busy waterways by using ais data}.
\newblock \bibinfo{journal}{Ocean Engineering} \bibinfo{volume}{37}, \bibinfo{pages}{483--490}.
\bibitem[{Mousavi et~al.(2016)Mousavi, Schukat and Howley}]{mousavi2016deep}
\bibinfo{author}{Mousavi, S.S.}, \bibinfo{author}{Schukat, M.}, \bibinfo{author}{Howley, E.}, \bibinfo{year}{2016}.
\newblock \bibinfo{title}{Deep reinforcement learning: an overview}, in: \bibinfo{booktitle}{Proceedings of SAI Intelligent Systems Conference}, \bibinfo{organization}{Springer}. pp. \bibinfo{pages}{426--440}.
\bibitem[{Nassif et~al.(2019)Nassif, Shahin, Attili, Azzeh and Shaalan}]{nassif2019speech}
\bibinfo{author}{Nassif, A.B.}, \bibinfo{author}{Shahin, I.}, \bibinfo{author}{Attili, I.}, \bibinfo{author}{Azzeh, M.}, \bibinfo{author}{Shaalan, K.}, \bibinfo{year}{2019}.
\newblock \bibinfo{title}{Speech recognition using deep neural networks: A systematic review}.
\newblock \bibinfo{journal}{IEEE Access} \bibinfo{volume}{7}, \bibinfo{pages}{19143--19165}.
\bibitem[{Negenborn et~al.(2023)Negenborn, Goerlandt, Johansen, Slaets, Valdez~Banda, Vanelslander and Ventikos}]{negenborn2023autonomous}
\bibinfo{author}{Negenborn, R.R.}, \bibinfo{author}{Goerlandt, F.}, \bibinfo{author}{Johansen, T.A.}, \bibinfo{author}{Slaets, P.}, \bibinfo{author}{Valdez~Banda, O.A.}, \bibinfo{author}{Vanelslander, T.}, \bibinfo{author}{Ventikos, N.P.}, \bibinfo{year}{2023}.
\newblock \bibinfo{title}{Autonomous ships are on the horizon: here’s what we need to know}.
\newblock \bibinfo{journal}{Nature} \bibinfo{volume}{615}, \bibinfo{pages}{30--33}.
\bibitem[{Ni et~al.(2021)Ni, Eysenbach and Salakhutdinov}]{ni2021recurrent}
\bibinfo{author}{Ni, T.}, \bibinfo{author}{Eysenbach, B.}, \bibinfo{author}{Salakhutdinov, R.}, \bibinfo{year}{2021}.
\newblock \bibinfo{title}{Recurrent model-free {RL} is a strong baseline for many {POMDPs}}.
\newblock \bibinfo{journal}{arXiv preprint arXiv:2110.05038} .
\bibitem[{Ozturk and Cicek(2019)}]{ozturk2019individual}
\bibinfo{author}{Ozturk, U.}, \bibinfo{author}{Cicek, K.}, \bibinfo{year}{2019}.
\newblock \bibinfo{title}{Individual collision risk assessment in ship navigation: A systematic literature review}.
\newblock \bibinfo{journal}{Ocean Engineering} \bibinfo{volume}{180}, \bibinfo{pages}{130--143}.
\bibitem[{Paielli and Erzberger(1997)}]{paielli1997conflict}
\bibinfo{author}{Paielli, R.A.}, \bibinfo{author}{Erzberger, H.}, \bibinfo{year}{1997}.
\newblock \bibinfo{title}{Conflict probability estimation for free flight}.
\newblock \bibinfo{journal}{Journal of Guidance, Control, and Dynamics} \bibinfo{volume}{20}, \bibinfo{pages}{588--596}.
\bibitem[{Pandey et~al.(2017)Pandey, Pandey and Parhi}]{pandey2017mobile}
\bibinfo{author}{Pandey, A.}, \bibinfo{author}{Pandey, S.}, \bibinfo{author}{Parhi, D.}, \bibinfo{year}{2017}.
\newblock \bibinfo{title}{Mobile robot navigation and obstacle avoidance techniques: A review}.
\newblock \bibinfo{journal}{International Robotics \& Automation Journal} \bibinfo{volume}{2}, \bibinfo{pages}{96--105}.
\bibitem[{Patle et~al.(2019)Patle, Pandey, Parhi, Jagadeesh et~al.}]{patle2019review}
\bibinfo{author}{Patle, B.}, \bibinfo{author}{Pandey, A.}, \bibinfo{author}{Parhi, D.}, \bibinfo{author}{Jagadeesh, A.}, et~al., \bibinfo{year}{2019}.
\newblock \bibinfo{title}{A review: On path planning strategies for navigation of mobile robot}.
\newblock \bibinfo{journal}{Defence Technology} \bibinfo{volume}{15}, \bibinfo{pages}{582--606}.
\bibitem[{Paulig and Okhrin(2024)}]{paulig2024open}
\bibinfo{author}{Paulig, N.}, \bibinfo{author}{Okhrin, O.}, \bibinfo{year}{2024}.
\newblock \bibinfo{title}{An open-source framework for data-driven trajectory extraction from {AIS} data--the $\alpha$-method}.
\newblock \bibinfo{journal}{arXiv preprint arXiv:2407.04402} .
\bibitem[{Polvara et~al.(2018)Polvara, Sharma, Wan, Manning and Sutton}]{polvara2018obstacle}
\bibinfo{author}{Polvara, R.}, \bibinfo{author}{Sharma, S.}, \bibinfo{author}{Wan, J.}, \bibinfo{author}{Manning, A.}, \bibinfo{author}{Sutton, R.}, \bibinfo{year}{2018}.
\newblock \bibinfo{title}{Obstacle avoidance approaches for autonomous navigation of unmanned surface vehicles}.
\newblock \bibinfo{journal}{The Journal of Navigation} \bibinfo{volume}{71}, \bibinfo{pages}{241--256}.
\bibitem[{Puterman(1994)}]{puterman1994markov}
\bibinfo{author}{Puterman, M.L.}, \bibinfo{year}{1994}.
\newblock \bibinfo{title}{Markov Decision Processes: Discrete Stochastic Dynamic Programming}.
\newblock \bibinfo{publisher}{John Wiley \& Sons}.
\bibitem[{Reif and Sharir(1994)}]{reif1994motion}
\bibinfo{author}{Reif, J.}, \bibinfo{author}{Sharir, M.}, \bibinfo{year}{1994}.
\newblock \bibinfo{title}{Motion planning in the presence of moving obstacles}.
\newblock \bibinfo{journal}{Journal of the ACM} \bibinfo{volume}{41}, \bibinfo{pages}{764--790}.
\bibitem[{Ribeiro et~al.(2020)Ribeiro, Ellerbroek and Hoekstra}]{ribeiro2020review}
\bibinfo{author}{Ribeiro, M.}, \bibinfo{author}{Ellerbroek, J.}, \bibinfo{author}{Hoekstra, J.}, \bibinfo{year}{2020}.
\newblock \bibinfo{title}{Review of conflict resolution methods for manned and unmanned aviation}.
\newblock \bibinfo{journal}{Aerospace} \bibinfo{volume}{7}, \bibinfo{pages}{79}.
\bibitem[{Roghair et~al.(2022)Roghair, Niaraki, Ko and Jannesari}]{roghair2022vision}
\bibinfo{author}{Roghair, J.}, \bibinfo{author}{Niaraki, A.}, \bibinfo{author}{Ko, K.}, \bibinfo{author}{Jannesari, A.}, \bibinfo{year}{2022}.
\newblock \bibinfo{title}{A vision based deep reinforcement learning algorithm for uav obstacle avoidance}, in: \bibinfo{booktitle}{Intelligent Systems and Applications: Intelligent Systems Conference}, \bibinfo{organization}{Springer}. pp. \bibinfo{pages}{115--128}.
\bibitem[{Sauer et~al.(2018)Sauer, Savinov and Geiger}]{sauer2018conditional}
\bibinfo{author}{Sauer, A.}, \bibinfo{author}{Savinov, N.}, \bibinfo{author}{Geiger, A.}, \bibinfo{year}{2018}.
\newblock \bibinfo{title}{Conditional affordance learning for driving in urban environments}, in: \bibinfo{booktitle}{Conference on Robot Learning}, \bibinfo{organization}{PMLR}. pp. \bibinfo{pages}{237--252}.
\bibitem[{Shen et~al.(2019)Shen, Hashimoto, Matsuda, Taniguchi, Terada and Guo}]{shen2019automatic}
\bibinfo{author}{Shen, H.}, \bibinfo{author}{Hashimoto, H.}, \bibinfo{author}{Matsuda, A.}, \bibinfo{author}{Taniguchi, Y.}, \bibinfo{author}{Terada, D.}, \bibinfo{author}{Guo, C.}, \bibinfo{year}{2019}.
\newblock \bibinfo{title}{Automatic collision avoidance of multiple ships based on deep q-learning}.
\newblock \bibinfo{journal}{Applied Ocean Research} \bibinfo{volume}{86}, \bibinfo{pages}{268--288}.
\bibitem[{Silver et~al.(2018)Silver, Hubert, Schrittwieser, Antonoglou, Lai, Guez, Lanctot, Sifre, Kumaran, Graepel et~al.}]{silver2018general}
\bibinfo{author}{Silver, D.}, \bibinfo{author}{Hubert, T.}, \bibinfo{author}{Schrittwieser, J.}, \bibinfo{author}{Antonoglou, I.}, \bibinfo{author}{Lai, M.}, \bibinfo{author}{Guez, A.}, \bibinfo{author}{Lanctot, M.}, \bibinfo{author}{Sifre, L.}, \bibinfo{author}{Kumaran, D.}, \bibinfo{author}{Graepel, T.}, et~al., \bibinfo{year}{2018}.
\newblock \bibinfo{title}{A general reinforcement learning algorithm that masters chess, shogi, and go through self-play}.
\newblock \bibinfo{journal}{Science} \bibinfo{volume}{362}, \bibinfo{pages}{1140--1144}.
\bibitem[{Simmons(1996)}]{simmons1996curvature}
\bibinfo{author}{Simmons, R.}, \bibinfo{year}{1996}.
\newblock \bibinfo{title}{The curvature-velocity method for local obstacle avoidance}, in: \bibinfo{booktitle}{International Conference on Robotics and Automation}, \bibinfo{organization}{IEEE}. pp. \bibinfo{pages}{3375--3382}.
\bibitem[{Su{\'a}rez-Varela et~al.(2019)Su{\'a}rez-Varela, Mestres, Yu, Kuang, Feng, Barlet-Ros and Cabellos-Aparicio}]{suarez2019feature}
\bibinfo{author}{Su{\'a}rez-Varela, J.}, \bibinfo{author}{Mestres, A.}, \bibinfo{author}{Yu, J.}, \bibinfo{author}{Kuang, L.}, \bibinfo{author}{Feng, H.}, \bibinfo{author}{Barlet-Ros, P.}, \bibinfo{author}{Cabellos-Aparicio, A.}, \bibinfo{year}{2019}.
\newblock \bibinfo{title}{Feature engineering for deep reinforcement learning based routing}, in: \bibinfo{booktitle}{International Conference on Communications}, \bibinfo{organization}{IEEE}. pp. \bibinfo{pages}{1--6}.
\bibitem[{Sutton and Barto(2018)}]{sutton2018reinforcement}
\bibinfo{author}{Sutton, R.S.}, \bibinfo{author}{Barto, A.G.}, \bibinfo{year}{2018}.
\newblock \bibinfo{title}{Reinforcement Learning: An Introduction}.
\newblock \bibinfo{publisher}{Cambridge: The MIT Press}.
\bibitem[{Szlapczynski and Szlapczynska(2017)}]{szlapczynski2017review}
\bibinfo{author}{Szlapczynski, R.}, \bibinfo{author}{Szlapczynska, J.}, \bibinfo{year}{2017}.
\newblock \bibinfo{title}{Review of ship safety domains: Models and applications}.
\newblock \bibinfo{journal}{Ocean Engineering} \bibinfo{volume}{145}, \bibinfo{pages}{277--289}.
\bibitem[{Tan et~al.(2020)Tan, Huang, Tan and Teo}]{tan2020three}
\bibinfo{author}{Tan, C.Y.}, \bibinfo{author}{Huang, S.}, \bibinfo{author}{Tan, K.K.}, \bibinfo{author}{Teo, R.S.H.}, \bibinfo{year}{2020}.
\newblock \bibinfo{title}{Three dimensional collision avoidance for multi unmanned aerial vehicles using velocity obstacle}.
\newblock \bibinfo{journal}{Journal of Intelligent \& Robotic Systems} \bibinfo{volume}{97}, \bibinfo{pages}{227--248}.
\bibitem[{Tengesdal et~al.(2020)Tengesdal, Johansen and Brekke}]{tengesdal2020risk}
\bibinfo{author}{Tengesdal, T.}, \bibinfo{author}{Johansen, T.A.}, \bibinfo{author}{Brekke, E.}, \bibinfo{year}{2020}.
\newblock \bibinfo{title}{Risk-based autonomous maritime collision avoidance considering obstacle intentions}, in: \bibinfo{booktitle}{International Conference on Information Fusion}, \bibinfo{organization}{IEEE}. pp. \bibinfo{pages}{1--8}.
\bibitem[{Tengesdal et~al.(2021)Tengesdal, Johansen and Brekke}]{tengesdal2021ship}
\bibinfo{author}{Tengesdal, T.}, \bibinfo{author}{Johansen, T.A.}, \bibinfo{author}{Brekke, E.F.}, \bibinfo{year}{2021}.
\newblock \bibinfo{title}{Ship collision avoidance utilizing the cross-entropy method for collision risk assessment}.
\newblock \bibinfo{journal}{IEEE Transactions on Intelligent Transportation Systems} \bibinfo{volume}{23}, \bibinfo{pages}{11148--11161}.
\bibitem[{Treiber and Kesting(2013)}]{Treiber2013}
\bibinfo{author}{Treiber, M.}, \bibinfo{author}{Kesting, A.}, \bibinfo{year}{2013}.
\newblock \bibinfo{title}{Traffic Flow Dynamics: Data, Models and Simulation}.
\newblock \bibinfo{publisher}{Springer-Verlag Berlin Heidelberg}.
\bibitem[{Tsay(2010)}]{tsay2005analysis}
\bibinfo{author}{Tsay, R.S.}, \bibinfo{year}{2010}.
\newblock \bibinfo{title}{Analysis of Financial Time Series}.
\newblock \bibinfo{publisher}{New Jersey: John Wiley \& Sons}.
\bibitem[{Vagale et~al.(2021)Vagale, Oucheikh, Bye, Osen and Fossen}]{vagale2021path}
\bibinfo{author}{Vagale, A.}, \bibinfo{author}{Oucheikh, R.}, \bibinfo{author}{Bye, R.T.}, \bibinfo{author}{Osen, O.L.}, \bibinfo{author}{Fossen, T.I.}, \bibinfo{year}{2021}.
\newblock \bibinfo{title}{Path planning and collision avoidance for autonomous surface vehicles {I}: a review}.
\newblock \bibinfo{journal}{Journal of Marine Science and Technology} \bibinfo{volume}{26}, \bibinfo{pages}{1292–1306}.
\bibitem[{Van~Hasselt et~al.(2016)Van~Hasselt, Guez and Silver}]{van2016deep}
\bibinfo{author}{Van~Hasselt, H.}, \bibinfo{author}{Guez, A.}, \bibinfo{author}{Silver, D.}, \bibinfo{year}{2016}.
\newblock \bibinfo{title}{Deep reinforcement learning with double q-learning}, in: \bibinfo{booktitle}{AAAI Conference on Artificial Intelligence}.
\bibitem[{Vinyals et~al.(2019)Vinyals, Babuschkin, Czarnecki, Mathieu, Dudzik, Chung, Choi, Powell, Ewalds, Georgiev et~al.}]{vinyals2019grandmaster}
\bibinfo{author}{Vinyals, O.}, \bibinfo{author}{Babuschkin, I.}, \bibinfo{author}{Czarnecki, W.M.}, \bibinfo{author}{Mathieu, M.}, \bibinfo{author}{Dudzik, A.}, \bibinfo{author}{Chung, J.}, \bibinfo{author}{Choi, D.H.}, \bibinfo{author}{Powell, R.}, \bibinfo{author}{Ewalds, T.}, \bibinfo{author}{Georgiev, P.}, et~al., \bibinfo{year}{2019}.
\newblock \bibinfo{title}{Grandmaster level in starcraft {II} using multi-agent reinforcement learning}.
\newblock \bibinfo{journal}{Nature} \bibinfo{volume}{575}, \bibinfo{pages}{350--354}.
\bibitem[{Waltz and Okhrin(2023)}]{WaltzOkhrin2022COLREG}
\bibinfo{author}{Waltz, M.}, \bibinfo{author}{Okhrin, O.}, \bibinfo{year}{2023}.
\newblock \bibinfo{title}{Spatial--temporal recurrent reinforcement learning for autonomous ships}.
\newblock \bibinfo{journal}{Neural Networks} \bibinfo{volume}{165}, \bibinfo{pages}{634--653}.
\bibitem[{Waltz et~al.(2023)Waltz, Paulig and Okhrin}]{waltz20232}
\bibinfo{author}{Waltz, M.}, \bibinfo{author}{Paulig, N.}, \bibinfo{author}{Okhrin, O.}, \bibinfo{year}{2023}.
\newblock \bibinfo{title}{2-level reinforcement learning for ships on inland waterways}.
\newblock \bibinfo{journal}{arXiv preprint arXiv:2307.16769} .
\bibitem[{Wang et~al.(2019)Wang, Wang, Shen and Zhang}]{wang2019autonomous}
\bibinfo{author}{Wang, C.}, \bibinfo{author}{Wang, J.}, \bibinfo{author}{Shen, Y.}, \bibinfo{author}{Zhang, X.}, \bibinfo{year}{2019}.
\newblock \bibinfo{title}{Autonomous navigation of {UAVs} in large-scale complex environments: A deep reinforcement learning approach}.
\newblock \bibinfo{journal}{IEEE Transactions on Vehicular Technology} \bibinfo{volume}{68}, \bibinfo{pages}{2124--2136}.
\bibitem[{Watkins and Dayan(1992)}]{watkins1992q}
\bibinfo{author}{Watkins, C.J.}, \bibinfo{author}{Dayan, P.}, \bibinfo{year}{1992}.
\newblock \bibinfo{title}{Q-learning}.
\newblock \bibinfo{journal}{Machine Learning} \bibinfo{volume}{8}, \bibinfo{pages}{279--292}.
\bibitem[{Xu et~al.(2020)Xu, Zhang, Jiang, Liu and Cheng}]{xu2020collision}
\bibinfo{author}{Xu, T.}, \bibinfo{author}{Zhang, S.}, \bibinfo{author}{Jiang, Z.}, \bibinfo{author}{Liu, Z.}, \bibinfo{author}{Cheng, H.}, \bibinfo{year}{2020}.
\newblock \bibinfo{title}{Collision avoidance of high-speed obstacles for mobile robots via maximum-speed aware velocity obstacle method}.
\newblock \bibinfo{journal}{IEEE Access} \bibinfo{volume}{8}, \bibinfo{pages}{138493--138507}.
\bibitem[{Xu et~al.(2022)Xu, Lu, Liu, Cai and Zhang}]{xu2022colregs}
\bibinfo{author}{Xu, X.}, \bibinfo{author}{Lu, Y.}, \bibinfo{author}{Liu, G.}, \bibinfo{author}{Cai, P.}, \bibinfo{author}{Zhang, W.}, \bibinfo{year}{2022}.
\newblock \bibinfo{title}{{COLREGs}-abiding hybrid collision avoidance algorithm based on deep reinforcement learning for {USVs}}.
\newblock \bibinfo{journal}{Ocean Engineering} \bibinfo{volume}{247}, \bibinfo{pages}{110749}.
\bibitem[{Xue et~al.(2019)Xue, Li, Zhang and Yan}]{xue2019deep}
\bibinfo{author}{Xue, X.}, \bibinfo{author}{Li, Z.}, \bibinfo{author}{Zhang, D.}, \bibinfo{author}{Yan, Y.}, \bibinfo{year}{2019}.
\newblock \bibinfo{title}{A deep reinforcement learning method for mobile robot collision avoidance based on double {DQN}}, in: \bibinfo{booktitle}{International Symposium on Industrial Electronics}, \bibinfo{organization}{IEEE}. pp. \bibinfo{pages}{2131--2136}.
\bibitem[{Yang et~al.(2019)Yang, Wu, Wang, Jia and Li}]{yang2019big}
\bibinfo{author}{Yang, D.}, \bibinfo{author}{Wu, L.}, \bibinfo{author}{Wang, S.}, \bibinfo{author}{Jia, H.}, \bibinfo{author}{Li, K.X.}, \bibinfo{year}{2019}.
\newblock \bibinfo{title}{How big data enriches maritime research--a critical review of {Automatic Identification System} ({AIS}) data applications}.
\newblock \bibinfo{journal}{Transport Reviews} \bibinfo{volume}{39}, \bibinfo{pages}{755--773}.
\bibitem[{Yang et~al.(2004)Yang, Yang, Kuchar and Feron}]{yang2004real}
\bibinfo{author}{Yang, L.}, \bibinfo{author}{Yang, J.H.}, \bibinfo{author}{Kuchar, J.}, \bibinfo{author}{Feron, E.}, \bibinfo{year}{2004}.
\newblock \bibinfo{title}{A real-time monte carlo implementation for computing probability of conflict}, in: \bibinfo{booktitle}{AIAA Guidance, Navigation, and Control Conference and Exhibit}, p. \bibinfo{pages}{4876}.
\bibitem[{Yu et~al.(2022)Yu, Teixeira, Liu and Soares}]{yu2022framework}
\bibinfo{author}{Yu, Q.}, \bibinfo{author}{Teixeira, A.}, \bibinfo{author}{Liu, K.}, \bibinfo{author}{Soares, C.G.}, \bibinfo{year}{2022}.
\newblock \bibinfo{title}{Framework and application of multi-criteria ship collision risk assessment}.
\newblock \bibinfo{journal}{Ocean Engineering} \bibinfo{volume}{250}, \bibinfo{pages}{111006}.
\bibitem[{Yun et~al.(2024)Yun, Dai, An, Zhang and Shang}]{yun2024doubly}
\bibinfo{author}{Yun, Y.}, \bibinfo{author}{Dai, H.}, \bibinfo{author}{An, R.}, \bibinfo{author}{Zhang, Y.}, \bibinfo{author}{Shang, X.}, \bibinfo{year}{2024}.
\newblock \bibinfo{title}{Doubly constrained offline reinforcement learning for learning path recommendation}.
\newblock \bibinfo{journal}{Knowledge-Based Systems} \bibinfo{volume}{284}, \bibinfo{pages}{111242}.
\bibitem[{Zhang et~al.(2020)Zhang, Liu, Ng and Low}]{zhang2020collision}
\bibinfo{author}{Zhang, N.}, \bibinfo{author}{Liu, H.}, \bibinfo{author}{Ng, B.F.}, \bibinfo{author}{Low, K.H.}, \bibinfo{year}{2020}.
\newblock \bibinfo{title}{Collision probability between intruding drone and commercial aircraft in airport restricted area based on collision-course trajectory planning}.
\newblock \bibinfo{journal}{Transportation Research Part C: Emerging Technologies} \bibinfo{volume}{120}, \bibinfo{pages}{102736}.
\bibitem[{Zhao and Liu(2021)}]{zhao2021physics}
\bibinfo{author}{Zhao, P.}, \bibinfo{author}{Liu, Y.}, \bibinfo{year}{2021}.
\newblock \bibinfo{title}{Physics informed deep reinforcement learning for aircraft conflict resolution}.
\newblock \bibinfo{journal}{IEEE Transactions on Intelligent Transportation Systems} \bibinfo{volume}{23}, \bibinfo{pages}{8288--8301}.
\bibitem[{Zheng et~al.(2020)Zheng, Chen, Jiang and Qiao}]{zheng2020svm}
\bibinfo{author}{Zheng, K.}, \bibinfo{author}{Chen, Y.}, \bibinfo{author}{Jiang, Y.}, \bibinfo{author}{Qiao, S.}, \bibinfo{year}{2020}.
\newblock \bibinfo{title}{A {SVM} based ship collision risk assessment algorithm}.
\newblock \bibinfo{journal}{Ocean Engineering} \bibinfo{volume}{202}, \bibinfo{pages}{107062}.
\bibitem[{Zhou et~al.(2021)Zhou, Liu, Pourpanah, Zeng and Wang}]{zhou2021survey}
\bibinfo{author}{Zhou, X.}, \bibinfo{author}{Liu, H.}, \bibinfo{author}{Pourpanah, F.}, \bibinfo{author}{Zeng, T.}, \bibinfo{author}{Wang, X.}, \bibinfo{year}{2021}.
\newblock \bibinfo{title}{A survey on epistemic (model) uncertainty in supervised learning: Recent advances and applications}.
\newblock \bibinfo{journal}{Neurocomputing} \bibinfo{volume}{489}, \bibinfo{pages}{449--465}.
\bibitem[{Zou et~al.(2021)Zou, Zhang, Zhong, Liu and Feng}]{zou2021collision}
\bibinfo{author}{Zou, Y.}, \bibinfo{author}{Zhang, H.}, \bibinfo{author}{Zhong, G.}, \bibinfo{author}{Liu, H.}, \bibinfo{author}{Feng, D.}, \bibinfo{year}{2021}.
\newblock \bibinfo{title}{Collision probability estimation for small unmanned aircraft systems}.
\newblock \bibinfo{journal}{Reliability Engineering \& System Safety} \bibinfo{volume}{213}, \bibinfo{pages}{107619}.

\end{thebibliography}

\newpage
\appendix

\setcounter{table}{0}
\setcounter{figure}{0}
\section{RL algorithm hyperparameters}
\begin{table}[H]    
    \centering
    \begin{tabular}{l|l}
    Hyperparameter & Value\\
    \toprule
        Discount factor  &  0.99 \\
        Batch size  & 32\\
        Loss function & Mean squared error\\
        Replay buffer size  & $10^6$ \\
        Learning rate actor  & $0.0001$ \\ 
        Learning rate critic  & $0.0001$ \\ 
        Optimizer & Adam \\
        Number of units per hidden layer & 128\\
        \midrule
        Target noise\textsuperscript{\textdagger} & 0.2\\
        Target noise clip\textsuperscript{\textdagger} & 0.5\\
        Policy update delay\textsuperscript{\textdagger} & 2\\
        \midrule
        Initial temperature* & 0.2\\
        Learning rate temperature* & 0.0001\\
    \end{tabular}
    \caption{List of hyperparameters of the RL algorithms. Parameters with a \textsuperscript{\textdagger} are solely used by the LSTM-TD3, whereas * refers to LSTM-SAC.}\label{tab:hyperparams_RL}
\end{table}

\setcounter{table}{0}
\setcounter{figure}{0}
\section{Environment details}\label{appendix:env_details}
At the beginning of a training episode, we initialize the agent's dynamics to zero, except the longitudinal speed $\Dot{x}_{\rm agent}$, that is sampled uniformly at random from the interval $[\unit[1]{m/s},v_{\rm max}]$.
The Euler and ballistic methods are used to update the agent's lateral speed and the positions for the agent and obstacles at time step $t+1$ \citep{Treiber2013}. Exemplary for the agent, we have:
\begin{align}
	\Dot{y}_{t+1, \rm agent} &= \Dot{y}_{t, \rm agent} +\Ddot{y}_{t+1, \rm agent}\Delta t,\\
	\Vec{p}_{t+1, \rm agent} &= \Vec{p}_{t, \rm agent} + \frac{\Dot{\Vec{p}}_{t, \rm agent} + \Dot{\Vec{p}}_{t+1, \rm agent}}{2} \Delta t,
\end{align}
with $\Delta t = \unit[5]{s}$ corresponding to the simulation step size. One episode is defined as $N_{\rm steps} = 500$ steps of updating. 

 For each obstacle, we define a passing rule. The set of obstacles that should only be passed, from the perspective of the agent, on the right side in the lateral direction is denoted as $\mathcal{M}_{\rm right} = \{1, \ldots, N_{\text{obstacle}} /2\}$. Consequently, the remaining obstacles should be passed left and are denoted $\mathcal{M}_{\rm left} = \{N_{\text{obstacle}}/2 + 1, \ldots, N_{\text{obstacle}}\}$.

\begin{figure}[ht]
	\centering
	\includegraphics[width=1\linewidth]{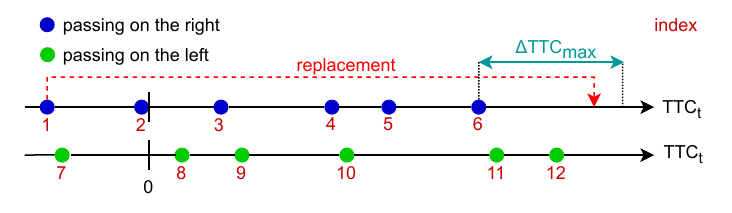}
	\caption{Replacement of an obstacle (obstacle $1$) since two obstacles with the same passing rule already passed the agent (negative time-to-collision). The obstacle's new $TTC_{t,1}$ is set uniformly at random in the time interval colored turquoise with the length of $\Delta TTC_{\rm max}$.}
	\label{fig:EnvB_TTC}
\end{figure}

Next, a procedure to update the trajectories of obstacles needs to be defined so that the agent has to face threatening obstacles constantly. Therefore, we replace obstacles that have already been passed. To specify this condition, we introduced the helper variable  $TTC_{t,i}$ as the agent's time-to-collision with an obstacle $i \in \mathcal{M}$ in the longitudinal direction at time step $t$, where the time-to-collision is computed based on the obstacle's linear trajectory part \eqref{eq:obstacle_traj_lin}. Negative values for $TTC_{t,i}$ relate to obstacles that have already passed the agent in the longitudinal direction. If for two obstacles $k, l \in \mathcal{M_{\rm right}}$ holds: $TTC_{t,k} < 0$, $TTC_{t,l} < 0$, and $TTC_{t,k} < TTC_{t,l}$, we replace obstacle $k$ as shown in Figure \ref{fig:EnvB_TTC}. Its new time-to-collision is randomly sampled from:
\begin{equation}
	\label{eq:EnvBTTC}
TTC_{t,k} \sim \mathcal{U}\left(\max_{j \in \mathcal{M_{\rm right}}}(TTC_{t,j}), \max_{j \in \mathcal{M_{\rm right}}}(TTC_{t,j}) + \Delta TTC_{\rm max}\right),
\end{equation}
where $\Delta TTC_{\rm max}$ is the maximal temporal distance for the new placement of an obstacle. This parameter affects the number of obstacles being passed in a certain time interval and is, therefore, a crucial design element of this environment. The same replacement procedure is applied for obstacles with passing rule 'left.'

\begin{figure}[ht]
	\centering
	\includegraphics[width=1\linewidth]{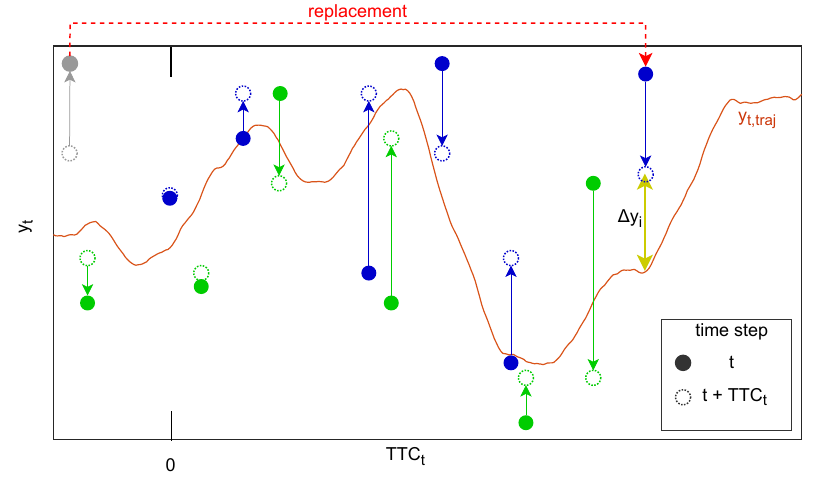}
	\caption{Replacement of an obstacle identical to the situation in Figure \ref{fig:EnvB_TTC} but with additional information about lateral positions of obstacles.}
	\label{fig:EnvB_technical}
\end{figure}

Having computed the new value for $TTC_{t,i}$ for a replaced obstacle $i \in \mathcal{M}$ at time step $t$, the new position and velocity need to be determined. First, we draw values for the constant velocity $\Dot{\Vec{p}}_{i} = (\Dot{x}_{i}, \Dot{y}_{i})^\top$ of the linear trajectory part from uniform distributions:
\vspace{-0.2cm}
\begin{align}	
	\Dot{x}_{i} &\sim \mathcal{U}(-v_{x,\rm max}, v_{x,\rm max}), \label{eq:EnvBxDot} \\		
	\Dot{y}_{i} &\sim \mathcal{U}(-v_{y,\rm max}, v_{y,\rm max}).
	\label{eq:EnvByDot}
\end{align} 
Second, the new longitudinal position can be set as follows:
\begin{equation}
	\label{eq:EnvBx}
	x_{t,i} = (\Dot{x}_{t,\rm agent} - \Dot{x}_{i})  TTC_{t,i} + x_{t,\rm agent}.
\end{equation}
Third, having the lateral speed of the replaced obstacle set, we generate the new lateral position with the help of a predefined, stochastic trajectory $y_{t,\rm traj}$, inspired by \cite{meyer2020taming}.~This lateral trajectory is computed at the beginning of an episode and is based on a smoothed AR(1) process, whose parameters reflect the kinematics of the agent. Figure \ref{fig:EnvB_technical} shows a replacement situation identical to Figure \ref{fig:EnvB_TTC} and illustrates how this trajectory is used to define the new lateral position for a replaced obstacle. One can think of this process as an exemplary trajectory the agent has to follow to avoid collisions with obstacles. In the following, we define the smoothed AR(1) process and give a detailed explanation of the replacement of an obstacle based on that process.

The AR(1) process is defined as:
\begin{equation} 
	B_{t+1}=\phi_{\rm traj} B_t + u_t, \quad \text{where} \quad u_t \sim \mathcal{N}(0, \sigma_{\rm traj}^2),
\end{equation}
with auto-regressive parameter $\phi_{\rm traj}$ and variance $\sigma_{\rm traj}^2$. The parameters have been designed to model a lateral trajectory the agent can approximately follow under acceleration and velocity constraints represented by $a_{y,\rm max}$ and $v_{\rm max}$. To reduce the noise, we exponentially smooth the AR(1) process:
\vspace{-0.3cm}
\begin{align}
	&y_{t,\rm traj}=
	\begin{cases}
		B_{0}, & \text{for } t = 0,\\
		\beta_{\rm traj} B_{t}+(1-\beta_{\rm traj}) y_{t-1,\rm traj}, & \text{for } t >0,
	\end{cases} 
\end{align}
where $\beta_{\rm traj}$ defines the smoothing factor. Based on this trajectory and having already computed $TTC_{t,i}$, $x_{t,i}$, $\Dot{x}_{i}$, and $\Dot{y}_{i}$ via (\ref{eq:EnvBTTC}), (\ref{eq:EnvBxDot}), (\ref{eq:EnvByDot}), and (\ref{eq:EnvBx}), one more step is needed to set the new lateral position $y_{t,i}$ for a replaced obstacle $i \in \mathcal{M}$ at time step $t$. 

We define $\Delta y_{i}$ as the absolute difference between an obstacle's lateral position $y_{t,i}$ and the defined trajectory $y_{t,\rm traj}$ when agent and obstacle are at the same longitudinal position ($TTC_{t,i} = 0$):
\begin{equation}
\Delta y_{i} = |y_{t,i} - y_{t,\rm traj}| \quad \text{for} \quad TTC_{t,i} = 0,
\end{equation}
shown yellow in Figure \ref{fig:EnvB_technical}.
To force our agent to move approximately along the trajectory $y_{t,\rm traj}$, the positional difference $\Delta y_{i}$ should be small, thus being another crucial design parameter to adjust the complexity of the environment. Every time an obstacle $i$ is replaced, the variable $\Delta y_{i}$ is sampled from a normal distribution:
\begin{equation}
\Delta y_{i}  \sim  \mathcal{N}(\mu_{\Delta y},\sigma_{\Delta y}^2),
\end{equation} 
and lower-bounded to $\Delta y_{\rm min}$:
\begin{equation}
\Delta y_{i} = \max(\Delta y_{\rm min}, \Delta y_{i}).		
\end{equation}
By changing the parameters $\mu_{\Delta y}$, $\sigma_{\Delta y}^2$, and $\Delta y_{\rm min}$, one can adjust how close the obstacles are coming to the trajectory $y_{t,\rm traj}$ when obstacle and agent are at the same longitudinal position. Table \ref{tab:EnvsParameters} contains the chosen values for those parameters. Finally, the lateral position for obstacles $i_{\rm R} \in \mathcal{M_{\rm right}}$ is computed via:
\begin{equation}
y_{t,i_{\rm R}} = y_{t,\rm traj} + \Delta y_{i_{\rm R}} - \Dot{y}_{i_{\rm R}}TTC_{t,i_{\rm R}},
\end{equation}
and for obstacles $i_{\rm L} \in \mathcal{M_{\rm left}}$ via:
\begin{equation}
y_{t,i_{\rm L}} = y_{t,\rm traj} - \Delta y_{i_{\rm L}} - \Dot{y}_{i_{\rm L}}TTC_{t,i_{\rm L}}.
\end{equation}
Figure \ref{fig:EnvB_technical} shows the final lateral position and time-to-collision for a replaced obstacle as a filled circle. The linear trajectory part is represented as a blue or green arrow. Having defined the linear trajectory part of an obstacle, we can now add the non-linear trajectory part based on \eqref{eq:obstacle_traj}.

\begin{table}
\caption{Environment parameters}
\label{tab:EnvsParameters} 
\begin{center}
	\begin{tabular}{ p{0.15\linewidth} |p{0.5\linewidth} |p{0.15\linewidth} } 
		Parameter & Description & Value   \\ \hline
             $N_{\text{obstacle}}$ & number of obstacles                             & 10 \\
		$a_{y,\rm max}$ 		& agent's maximum lateral acceleration  	    & $\unit[0.01]{m/s^2}$  \\
		$v_{\rm max}$ 		& agent's maximum speed  	    	& $\unit[5]{m/s}$   \\
		$\Delta t$              & simulation step size                          & $\unit[5]{s}$\\
           
	    $N_{\rm steps}$ & number of episode steps                               & 500 \\
		$p_{\rm scale}$			& scaling factor for observation  				& $\unit[3000]{m}$   \\
		$d^{\rm CPA}_{\rm scale}$			& scaling factor for observation  				& $\unit[400]{m}$   \\
		$t^{\rm CPA}_{\rm scale}$			& scaling factor for observation  				& $\unit[300]{s}$   \\
		$\Delta TTC_{\rm max}$	& maximal temporal distance for replacing an obstacle	& $\unit[300]{s}$   \\
		$\phi_{\rm stoch}$ 		        & AR(1) process	parameter		                	& $0.9$  \\
		$\phi_{\rm traj}$ 		        & AR(1) process	parameter		                	& $0.99$  \\
            $\sigma^2_{\rm stoch}$ 	& normal distribution variance      		    & $\unit[225]{m^2}$\\
            $\sigma^2_{\rm traj}$ 	& normal distribution variance      		    & $\unit[800]{m^2}$\\
            $\sigma^2_{\rm sin}$ 	& normal distribution variance      		    & $\unit[16]{m^2}$\\
		$\beta_{\rm stoch}$ 			& smoothing factor 									& $0.2$   \\
		$\beta_{\rm traj}$ 			& smoothing factor 									& $0.03$   \\
		$\mu_{\Delta y}$ 	& normal distribution mean							& $\unit[100]{m}$   \\
		$\sigma_{\Delta y}^2 $ & normal distribution variance					& $\unit[2500]{m^2}$   \\
		$\Delta y_{\rm min}$& minimum bound for $\Delta y$						& $\unit[40]{m}$   \\
		$A_{\rm sin}$ & periodic trajectory amplitude						& $\unit[15]{s}$   \\
		$T_{\rm sin}$ & periodic trajectory period						& $\unit[20]{s}$   \\
	\end{tabular}
\end{center}
\end{table}






\end{document}